\documentclass[11pt,a4paper]{article}
\usepackage[hyperref]{naaclhlt2019}
\usepackage{times}
\usepackage{latexsym}
\usepackage{amsmath}
\usepackage{amssymb}
\usepackage{url}
\usepackage{graphicx}
\usepackage{tikz}
\usepackage{subcaption}
\usepackage{xspace}
\usepackage{booktabs}
\usepackage{multirow}
\usepackage{chngcntr}
\usepackage{dsfont}
\aclfinalcopy %
\def\aclpaperid{1962} %

\title{
No Permanent Friends or Enemies:\\
Tracking 
Relationships between Nations 
from News }
\author{Xiaochuang Han \\
  College of Computing \\
  Georgia Institute of Technology \\
  Atlanta, GA \\
  {\tt xc@gatech.edu} \\\And
  Eunsol Choi \\
  Paul G. Allen School of CSE \\
  University of Washington\\
  Seattle, WA\\
  {\tt \small eunsol@cs.washington.edu} \\ \And
  Chenhao Tan \\
  Dept. of Computer Science \\
  Unversity of Colorado Boulder \\
  Boulder, CO \\
  {\tt chenhao@chenhaot.com} \\
  }

\date{}

\newif\ifcomment
\commenttrue
\ifcomment
    \newcommand{\chenhao}[1]{{\textbf{\color{red}{#1}}}}
    \newcommand{\eunsol}[1]{{\textbf{\color{blue}{[EC:#1]}}}}
    \newcommand{\han}[1]{{\textbf{\color{green}{#1}}}}
\else
    \newcommand{\chenhao}[1]{}
    \newcommand{\eunsol}[1]{}
    \newcommand{\han}[1]{}
\fi
\newcommand{\para}[1]{\paragraph{#1}}
\newcommand{\figref}[1]{Figure~\ref{#1}\xspace}

\newcommand{\rmn}{\emph{RMN}\xspace}
\newcommand{\modelname}{\emph{LARN}\xspace}
\newcommand{\articlecollection}{\mathcal{A}_{e_i,e_j}}

\begin{document}
\maketitle
\begin{abstract}
Understanding the dynamics of international politics is important yet challenging for civilians. In this work, we explore unsupervised neural models to infer relations between nations from news articles. We extend existing models by incorporating shallow linguistics information and propose a new automatic evaluation metric that aligns relationship dynamics with manually annotated key events. As understanding international relations requires carefully analyzing complex relationships, we conduct in-person human evaluations with three groups of participants. 
Overall, humans prefer the outputs of our model 
and give insightful feedback that suggests future directions for human-centered models.
Furthermore, our model reveals interesting regional differences in news coverage. For instance, with respect to US-China relations, Singaporean media focus more on ``strengthening'' and ``purchasing'', while US media focus more on ``criticizing'' and ``denouncing''. 

\end{abstract}

\section{Introduction}
\label{sec:introduction}

In the context of growing globalization~\citep{baylis2017globalization}, understanding complex international relations is increasingly relevant to our daily life. 
Yet this is a challenging task due to the inherently dynamic nature of international relations.
As Kissinger famously said, ``America has no permanent friends or enemies, only interests.''
Staying informed becomes even harder in the continuous streams of information from news outlets and social media.

This very availability of such information,
however, opens up exciting opportunities for natural language processing to support individuals in understanding international relations.
Supervised extraction has been incredibly useful at identifying pre-defined relations and events \cite{doddington2004automatic,mintz2009distant} but 
fails to capture emerging or complex information needs. 
Topic models and neural models have been proposed to explore relations between entities without supervision~\cite{o2013learning,Chaney2016DetectingAC,Iyyer:16}.
In particular,
\citet{Iyyer:16} introduces an unsupervised neural model for tracking 
relations between fictional characters, and this approach outperforms baselines from topic models and hidden Markov models.
In this work,
we 
incorporate linguistic insights into this model to track  relation dynamics between nations from news articles.%
Our model reconstructs textual information in the embedding space using relation embeddings, as proposed in \citet{Iyyer:16}.
We integrate simple yet effective linguistic insights:
verbal predicates often describe the relationship between entities,\footnote{We use entities, countries, and nations interchangeably in this work.} while nouns and proper nouns provide the context of this relationship.
For example, in ``U.S. denounces Russia for its interference in the 2016 election", \textit{denounce} describes the relation, and \textit{election} and \textit{interference} provide the context. 
We show that this intuition leads the model to discover relation descriptors that are easier to interpret and less noisy. %

Evaluating these exploratory 
models for subjective tasks poses a challenge as 
there are no gold labels.
Along with the model, we propose new approaches for evaluation. 
We introduce a quantitative metric which aligns pre-annotated key events with the temporal trends of relationships produced by the models.  
Since this task requires careful analysis of complex international relations,
we conduct in-person user studies with NLP researchers and undergraduate students recruited from political science and linguistics courses. 
Both quantitative evaluation and human evaluation indicate that our model better represents the dynamic relationships between nations than the prior model~\begin{NoHyper}\cite{Iyyer:16}\end{NoHyper}: 75.9\% of participants preferred our model for finding natural language words describing international relations and 85.5\% preferred temporal trends generated by our model.

Finally, we 
qualitatively explore the context of relations provided by 
an attention-based mechanism and demonstrate
a practical application of our model by studying regional differences in news coverage of relationships between two countries. 
We conclude with discussions on future directions for buildings models that can support individuals in navigating a large collection of news articles. 
Our code is available at \url{https://github.com/BoulderDS/LARN}.
\section{Data}

We start by introducing our dataset of news articles, the shallow linguistic information that we extract, and our annotation of key events.

\para{News article collection.}
Our dataset is derived from the NOW corpus, the largest corpus of news articles that is available in full-text format.\footnote{The dataset can be obtained from \url{https://www.corpusdata.org/now_corpus.asp}.}
The NOW corpus collects the news articles that are available on Google News in 20 English-speaking countries and thus include news articles that span a wide variety of topics, ranging from politics, to sports, to celebrity news from 23K media outlets.
In this work, we consider the news articles in recent years, i.e., from January 2016 to June 2018, to facilitate human evaluation.

We consider 12 nations (U.S., Russia, China, UK, Germany, Canada, France, India, Japan, Iran, Israel, and Syria) and the 66 nation pairs between them.
To identify mentions of each nation in news articles, we manually construct a set of aliases for each nation to cover common abbreviations and the names of political leaders (e.g., Trump, Putin). On average, each nation has 3.5 aliases. We then use these aliases to find sentences that contain a pair of nations under consideration and obtain 1.2M sentences associated with 634K articles.

\paragraph{Adding Shallow Linguistic Information.}
To incorporate shallow linguistic knowledge, we process the news article collection for each nation pairs to extract
(1) verbal predicates and (2) nouns and proper nouns from each sentence.
Specifically, we use a dependency parser to detect verbal predicates
and their subjects and objects, 
and only include predicates for which both subjects and objects were found. 
For sentences with such predicates, we find nouns and proper nouns using part of speech tags.
All data processing was done in spaCy~\citep{spacy2} and full details can be found in the appendix.

\para{Key events annotation.}
The main goal of our model is to support the exploration of international relations, which is very challenging to evaluate.
To derive quantitative evaluation measures and provide the basic context of international relations,
we %
manually identify key events over the 30 months for 8 most frequently mentioned nation pairs (i.e.
US-[China, Russia, UK, India, Canada, Japan, Syria], China-India) by reading through the top Google search results for each two countries and each month.\footnote{
This annotation is done by the first author and is thus inherently subjective.
We add a robustness check in the appendix based on another set of independently annotated key events by the third author. Automatic evaluation shows a similar trend holds for the two sets of annotations.}
We identified roughly five key events per nation pair.
For example, one key event for US-China relation is Chairman Xi's visit to the US in April 2017.
A complete list of key events is shown in the appendix.
Table~\ref{tab:data} summarizes the data statistics.

\begin{table}
\small
\begin{center}
\begin{tabular}{lr}
\toprule
Total \# of months considered & 30 \\
Total \# of articles & 7.7M \\
Total \# of articles with valid nation pairs & 634K\\\midrule
Avg. \# of articles per nation pair & 9.6K\\  %
Avg. \# of sentences per nation pair & 18.2K\\ %
Avg. \# of predicates per nation pair & 21.6K \\ %
Avg. \# of nouns \& proper nouns per nation pair & 201K \\  \midrule %
Avg. \# of key events per nation pair (8 pairs) & 4.9 \\ 
\bottomrule
\end{tabular} 
\end{center}
\caption{Data statistics.}
\label{tab:data}
\end{table}

\section{Model}
\label{sec:model}
In this section, we formally introduce our model that builds on Relation Modeling Network (\rmn)~\cite{Iyyer:16}.
Our main contribution is to integrate shallow linguistic information (i.e., verbal predicates and nouns/proper nouns) and identify the context of relations.

\subsection{Overview} %
The intuition behind our model follows RMN, i.e., inferring relation embeddings by reconstructing textual information in the embedding space.
Specifically, we learn a fixed set of relation embeddings and use a convex combination of these relation embeddings to reconstruct information in sentences that mention both entities.

Our main hypothesis is that relation information is often encoded in verbal predicates and we can obtain more interpretable and robust relations if we focus on predicates.
For each pair of entities, we extract information from the predicates in sentences with both entities and reconstruct 
these predicates using shared relation embeddings.
In addition, we use nouns to provide the context for the relations.
We refer to our model as Linguistically Aware Relationship Networks (\modelname).
We now formally define our problem.
The input of our model is a collection of news articles. For each entity pair $e_i, e_j$, we obtain a set of articles, $\articlecollection$, containing at least one sentence mentioning both entities.
We identify sentences where both $e_i$ and $e_j$ occur based on any alias associated with an entity (nation).
We extract all the verbal predicates  from these sentences in article $a \in \articlecollection$
as $\{p^{a, e_i, e_j}_1, \ldots, p^{a, e_i, e_j}_N\}$, as well as the proper nouns or nouns as $\{n^{a, e_i, e_j}_1, \ldots, n^{a, e_i, e_j}_M\}$.
Preprocessing details could be found in the appendix.
We then use GloVe 
embeddings to represent these words, i.e., $v_{p^{a, e_i, e_j}_k}$ for $p^{a, e_i, e_j}_k$ and $v_{n^{a, e_i, e_j}_k}$ for $n^{a, e_i, e_j}_k$~\cite{pennington2014glove}.
These word embeddings are static in the entire learning process.

Our model learns relation embeddings $\textbf{R} \in \mathbb{R}^{K \times d}$, where $K$ is a hyperparameter for the number of relations and $d$ is the dimension of %
the relation embedding as well as the word embedding.
Following \citet{Iyyer:16}, we obtain a list of natural language descriptors for each relation using the nearest neighbors of the relation embedding within the 500 most common predicates.%
\footnote{Refer to the appendix for a version that include all words. The relation descriptors from \rmn required a manual filtering step in \citet{Iyyer:16}, and become unintelligible without the 500 common words constraint. 
Note that our model produces intelligible descriptors even without the 500 most common predicates constraint.
} The model also provides (1) a probability distribution over relations for each article, (2) a probability distribution over nouns for each relation between each entity pair,
and (3) an embedding for each entity.

Figure~\ref{fig:model} describes the overall architecture. %
We will describe the construction of $v^a_{\text{label}}$ in \S\ref{subsec:model_predicates}, a reconstruction of $v^a_{\text{label}}$ through a weighted sum of relation embeddings in \S\ref{subsec:model_desc}, and finally the learning in \S\ref{subsec:model_obj}.

\definecolor{g-red}{HTML}{DB4437}
\definecolor{g-blue}{HTML}{4285F4}
\definecolor{g-green}{HTML}{0F9D58}
\definecolor{g-purple}{HTML}{9370DB}
\definecolor{g-yellow}{HTML}{F4B400}
\definecolor{g-orange}{HTML}{FF9800}
\definecolor{g-grey}{HTML}{9E9E9E}
\newcommand\layerbox[4]{
\draw[rounded corners] (#2, #3) rectangle (#2 + #1 * #4, #3 + #1 * 1);
}
\newcommand\layercolorbox[5][0.4] {
\draw[rounded corners, fill=#5] (#2, #3) rectangle (#2 + 0.6 * #4, #3 + 0.6 * 1);
\draw[-latex, line width=1pt] (#2+0.3,#3+0.6) -- (#2+0.3,#3+1.1);
}

\newcommand\layercomponent[5]{
\filldraw[fill=#5] (#2 + #1 * #4 - #1 * 0.5, #3 + #1 * 0.5) circle (#1 * 0.4);
}
\newcommand\layer[5][0.5] {
\layerbox{#1}{#2}{#3}{#4}
\foreach \x in {1, ..., #4}{
  \layercomponent{#1}{#2}{#3}{\x}{#5}
}
}

\newcommand\middletext[6] {
\node[anchor=east] at (#1, #2+0.22) {#5}; %
\layer{#1}{#2}{2}{#6};%
\draw[-latex, line width=1pt] (#1+0.5, #2+0.5) to (#1+#3, #2+0.8+#4); %
}

\newcommand\inputtext[6] {
\node[anchor=mid] at (#1+0.5, 0) {#4}; %
\node[anchor=mid] at (#1+0.5, -0.5) { #5}; %
\layer{#1}{0.3}{2}{#6};%
\draw[-latex, line width=1pt] (#1+0.5, 0.8) to (#1+#2, #3+0.8); 
}

\newcommand\datec[6] {
\node[anchor=mid] at (#1+0.5, 0) {#4}; %
\node[anchor=mid] at (#1+0.7, -0.6) {\footnotesize #5}; %
\layer{#1+0.2}{0.3}{1}{#6};%
}

\newcommand\matrixf[1]{
\layer{6.1}{5+#1}{2}{g-green};
\layer{6.1}{4.5+#1}{2}{g-green};
\layer{6.1}{4+#1}{2}{g-green};
\node[anchor=west] at (7.2,#1+4.5) {\textbf{R}: Relation embeddings};
}
\newcommand\sumnode[3] {
\layercolorbox{#1-0.3}{#2-0.3}{1}{#3};
\node at (#1, #2) {\textcolor{white}{+}};
}

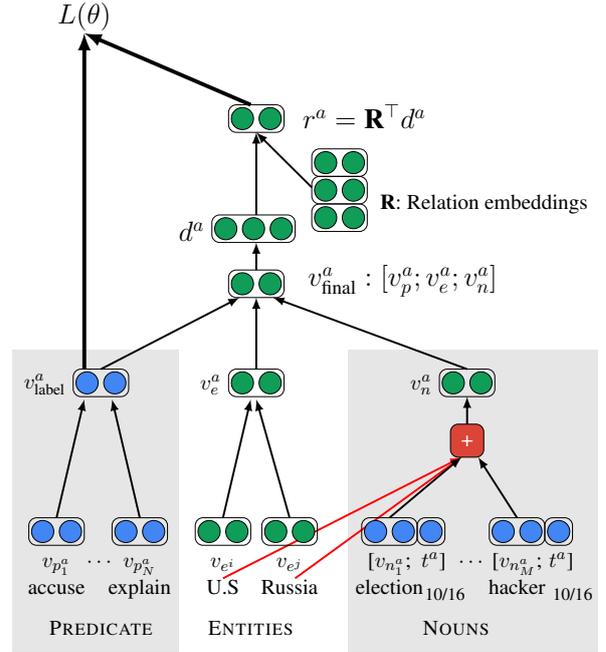
\begin{figure}[t!]
\begin{centering}
\scalebox{0.73} {
\begin{tikzpicture}

\node[rectangle, fill=black!10, minimum width=4.5cm,minimum height = 5.5cm] at (9,1.1) {};
\node[rectangle, fill=black!10, minimum width=3cm,minimum height = 5.5cm] at (2.2,1.1) {};
\draw[-latex, line width=1pt,color=red](4.5,-0.3) to (8.8,1.9); 
\draw[-latex, line width=1pt,color=red](5.8,-0.3) to (8.8,1.9);

\node[anchor=center] at (2.3, -1.2) {\textsc{Predicate}}; 
\inputtext{1}{1}{2.15}{$v_{p^a_1}$}{accuse}{g-blue}
\node[anchor=mid] at (2.3, 0) {$\cdots$}; 
\inputtext{2.5}{0}{2.15}{$v_{p^a_N}$}{explain}{g-blue}
\middletext{1.8}{3}{3}{1}{$v^a_{\text{label}}$}{g-blue}

\node[anchor=center] at (5, -1.2) {\textsc{Entities}}; 
\inputtext{4}{1}{2.15}{$v_{e^i}$}{U.S}{g-green}
\inputtext{5.2}{-0.1}{2.15}{$v_{e^j}$}{Russia}{g-green}
\middletext{4.6}{3}{0.5}{1}{$v^a_e$}{g-green}
\sumnode{8.9}{2.2}{g-red}

\node[anchor=center] at (8.7, -1.2) {\textsc{Nouns}}; 
\inputtext{7}{1.8}{1.1}{[$v_{n^a_1}$;}{election}{g-blue}
\datec{7.8}{0.4}{1.1}{$t^a$]}{10/16}{g-blue}
\node[anchor=mid] at (9, 0) {$\cdots$}; 
\inputtext{9.3}{-0.2}{1.1}{[$v_{n^a_M}$;}{hacker}{g-blue}
\datec{10.1}{-1}{1.1}{$t^a$]}{10/16}{g-blue}
\middletext{8.4}{3}{-3}{1}{$v^a_{n}$}{g-green}

\layer{4.6}{4.8}{2}{g-green}; %
\node[anchor=west] at (5.9, 5.1) {\Large $v^a_{\text{final}}:[v^a_{p};v^a_e;v^a_n]$};

\layer{4.3}{5.8}{3}{g-green}; %
\node[anchor=east] at (4.3, 6) {\Large $d^{a}$}; 

\draw[-latex, line width=1pt](5.1,5.3) to (5.1,5.8); %
\draw[-latex, line width=1pt](5.1,6.3) to (5.1,7.8); %
\draw[-latex, line width=1pt](6.1,6.8) to (5.1,7.8); %

\draw[-latex, line width=2pt, color=black](2.0,3.5) to (2.0,9.6);  %
\draw[-latex, line width=2pt](5,8.3) to (2.0,9.6); %

\layer{4.6}{7.8}{2}{g-green}; %
\node[anchor=west] at (5.8, 8.1) {\Large $r^a=\textbf{R}^ \top d^a$}; 

\node[anchor=north] at (2, 10.3) {\Large $L(\theta)$}; 
\matrixf{2}
\end{tikzpicture}}
\end{centering} 
\vspace{-10pt}
\caption{
    Model overview of an article with respect to a pair of nations.
   Our model approximates the predicates from the input article $v^a_{\text{label}}$ as a weighted sum of relation embeddings from \textbf{R}. The blue embeddings are fixed and the green ones are trained. 
   The red cell indicates our attention mechanism. The shaded area is our extension: we use only predicates as the label and provide the context of relations via nouns.}
    \label{fig:model}
\end{figure}

\subsection{Modeling Text with Predicates}\label{subsec:model_predicates}
We compute the representation for each article to be reconstructed as the sum of bag-of-words embeddings.
While \citet{Iyyer:16} considers all the words in a window in which both entities occur,
we only consider predicates in the sentences where both entities show up:
\begin{equation*}
    v^{a, e_i, e_j}_{\text{label}} = \sum_{k=1}^{N}v_{p^{a, e_i, e_j}_k}.
\end{equation*}
\noindent 
$v^{a, e_i, e_j}_{\text{label}}$ depends on both the news article and the entity pair.
In the rest of the paper, We omit $e_i, e_j$ in the superscript here for simplicity, i.e., $v^{a}_{\text{label}}$.

\subsection{Reconstructing Text with Relation Embeddings}\label{subsec:model_desc}
We represent each article as a weighted sum of 
relation embeddings.
Assuming $d^{a}$ 
represents 
a weight vector over the $K$ relation embeddings that sums to 1
for an article $a$ with respect to a pair of entities $(e_i, e_j)$, we obtain $r^a$ as follows:

\begin{equation*}
    r^{a} =  \textbf{R}^ \top d^{a}.
\end{equation*}

\noindent $d^{a}$ can also be thought of as a distribution over relations. 
This distribution over relations 
depends on 
entity pair ($v^a_{e}$), predicate information ($v^a_{p}$), and noun information ($v^a_{n}$). While \rmn simply takes the average of all words in the sentence, \modelname 
focuses on verbal predicates and nouns to capture our intuition that predicates describe the main relations, whereas nouns 
provide background information to explain the relations. 
We now describe these three components in details.

\para{Representing predicates and entity pair.}
We follow \rmn to construct embedding for each entity pair and for the predicates. The entity pair vector, $v^a_e$, simply adds the embedding of the two entities. The predicate vector, $v^a_p$, is equivalent to $v^a_{\text{label}}$ except for word dropout during training, i.e., 
setting $b^a_k$ to be 0 or 1 with a probability of 0.5. %
\begin{align*}
     v^a_{e} &= v_{e_i} + v_{e_j},\\
     v^a_{\text{p}} & = \sum_{k=1}^{N}b^a_k \cdot v_{p^a_k}.  %
\end{align*}

\para{Representing context with nouns.}
To understand relations between an entity pair in a sentence, nouns should be considered in addition to predicates. For example, 
{\em tariff} is indicative of the relation between US and China in ``Originally, Trump favoured the simple imposition of a tariff on products from selected countries, especially China and Mexico'', despite the seemingly positive predicate {\em favour}.
As nouns are much more common than predicates (see Table~\ref{tab:data}) and not all of them are meaningful for understanding international relations,
we employ a weighted sum of noun vectors.
We use an attention mechanism~\cite{Conneau2017SupervisedLO,Bahdanau} and consider each entity pair as a unique key to compute the attention weights, since the same noun can be interpreted differently across different entity pairs. To this end, we train an attention query embedding $q_{e_i,e_j}$ for each entity pair separately.
We further encode the temporal information by concatenating a one-hot vector $t^a$ that indicates the month when the article was published with the
noun representation $v_{n^a_k}$. 
This allows us to capture the shifts in a word's meaning over time.
\begin{align*}
    {h_{n^a_k}} &= \text{tanh}(W_{\text{proj}} \cdot [v_{n^a_k};t^a]),\\
    \alpha_{n^a_k} &= \dfrac{\text{exp}({h_{n^a_k}} \cdot q_{e_i, e_j})}{\sum_{k'=1}^{M} \text{exp}(h_{n^a_{k'}} \cdot q_{e_i, e_j})},\\
    v^a_{n} &= \sum_{k=1}^{M} \alpha_{n^a_k} h_{n^a_k}.
\end{align*}
Finally, we concatenate the three representations as the input to a feedforward network 
and pass to a softmax layer to create the weight vector $d^a$.
\begin{align*} 
    v_{\text{final}}^a &= \text{ReLU}(W_{\text{cat}} \cdot [v^a_{p}; v^a_{e}; v^a_{n}]),\\
     d^a &= \text{Softmax}(W_{\text{final}} \cdot v_{\text{final}}^a).
\end{align*}
This $d^a$ is multiplied with the descriptor matrix \textbf{R} to get the final representation $r^a$.
Different from \rmn, we do not consider temporal dependencies between time steps in our model
because it is important to understand sudden shifts in international relations rather than assuming that the relations slowly evolve.
We also found the temporal dependencies were not helpful empirically in our domain but rather computationally expensive.

\subsection{Learning Objective and Summary}\label{subsec:model_obj}

The reconstruction objective pushes $r^a$ to resemble $v^a_{label}$. Our formulation is identical to \rmn: the loss function consists of a contrastive max-margin loss term, $J$, and an auxiliary loss term, $X$, to encourage unique relation embeddings.
\begin{align*}
L(\theta) =& J(\theta) + \lambda X(\theta),\\
J(\theta) =& \sum_{a \in \articlecollection}^{} \sum_{v^a_{\text{label}'}\in \mathcal{N}} \max(0, \\
&\hspace{1em}1 - \dfrac{r_a \cdot v^a_{\text{label}}}{||r_a|| \cdot ||v^a_{\text{label}}||} + \dfrac{r_a \cdot v^a_{\text{label}'}}{||r_a|| \cdot ||v^a_{\text{label}'}||}),\\
X(\theta) =& \lVert RR^\top - I \rVert,
\end{align*}
\noindent where $v_{\text{label}}'$ is a randomly sampled negative example, $\mathcal{N}$ is a collection of them, and $\lambda$ is a hyperparameter for balancing two loss terms.

\section{Evaluation}

\begin{table*}[t]
\small
\begin{center}
\begin{tabular}{p{0.37\textwidth}r|p{0.37\textwidth}r}
\toprule
\modelname (Linguistically Aware Relationship Network) & Weight&\rmn (Relationship Modeling Network) & Weight  \\
\midrule
\shortstack[l]{denounce, undermine, condemn, punish, oppose} & 5.46\%&
\shortstack[l]{seem, thing, though, too, especially} & 11.79\%  \\  \shortstack[l]{leave,tell,ask,know,want} & 4.93\%&
\shortstack[l]{range, information, supply, infrastructure, value} & 9.42\%\\
\shortstack[l]{differ, indicate, affect, regard, determine} & 4.64\%&
\shortstack[l]{theresa, emmanuel, yemen, poland, lebanon} & 8.24\% \\
\shortstack[l]{strengthen, enhance, improve, develop, boost} & 4.56\%&
\shortstack[l]{negotiate, establish, impose, propose, negotiation} & 7.92\%\\
\shortstack[l]{hit, cut, end, kick, beat} & 4.05\%&
\shortstack[l]{terrorism, militant, terror, terrorist, condemn} & 7.54\% \\
\midrule
\shortstack[l]{buy, manufacture, use, brand, purchase} & 2.59\% & \shortstack[l]{propose, woman, represent, indians, spend} & 1.69e-09 \\
\shortstack[l]{win, defeat, beat, title, match} & 2.59\% & \shortstack[l]{de, defense, kremlin, cite, pressure} & 1.62e-09 \\
\shortstack[l]{receive, express, praise, send, acknowledge} & 2.58\% & \shortstack[l]{car, de, bomb, little, note} & 5.39e-10 \\
\shortstack[l]{offer, provide, deliver, feature, guarantee} & 2.46\% & \shortstack[l]{cost, cite, compare, co, m} & 2.94e-10 \\
\shortstack[l]{launch, announce, unveil, release, celebrate} & 2.36\% & 
\shortstack[l]{aid, hand, small, public, round} & 2.21e-10 \\
\bottomrule
\end{tabular} 
\end{center} \vspace{-1pt}
\caption{Relation descriptors of the most/least frequent five relations by \modelname and \rmn and their average weights in news articles.
The relations generated from \modelname are more semantically meaningful. %
}  
\label{tab:descriptor}
\end{table*}
\begin{figure*}
    
    \begin{subfigure}[t]{0.47\textwidth}
    \includegraphics[width=\textwidth]{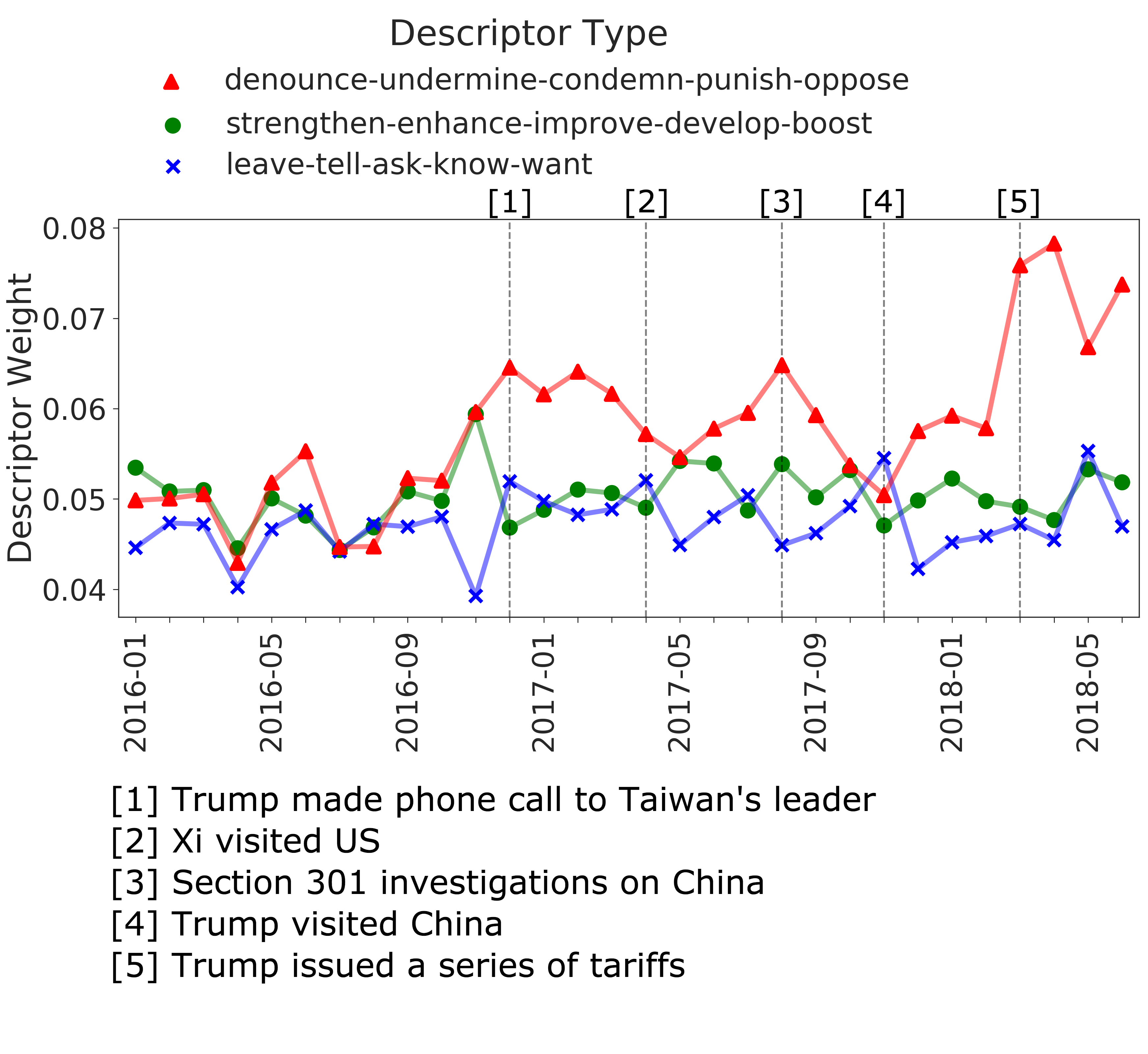}
    \caption{US-China's relation trends by \modelname.}
    \label{fig:trend_ours}
    \end{subfigure}
    \hfill
    \begin{subfigure}[t]{0.47\textwidth}
    \includegraphics[width=\textwidth]{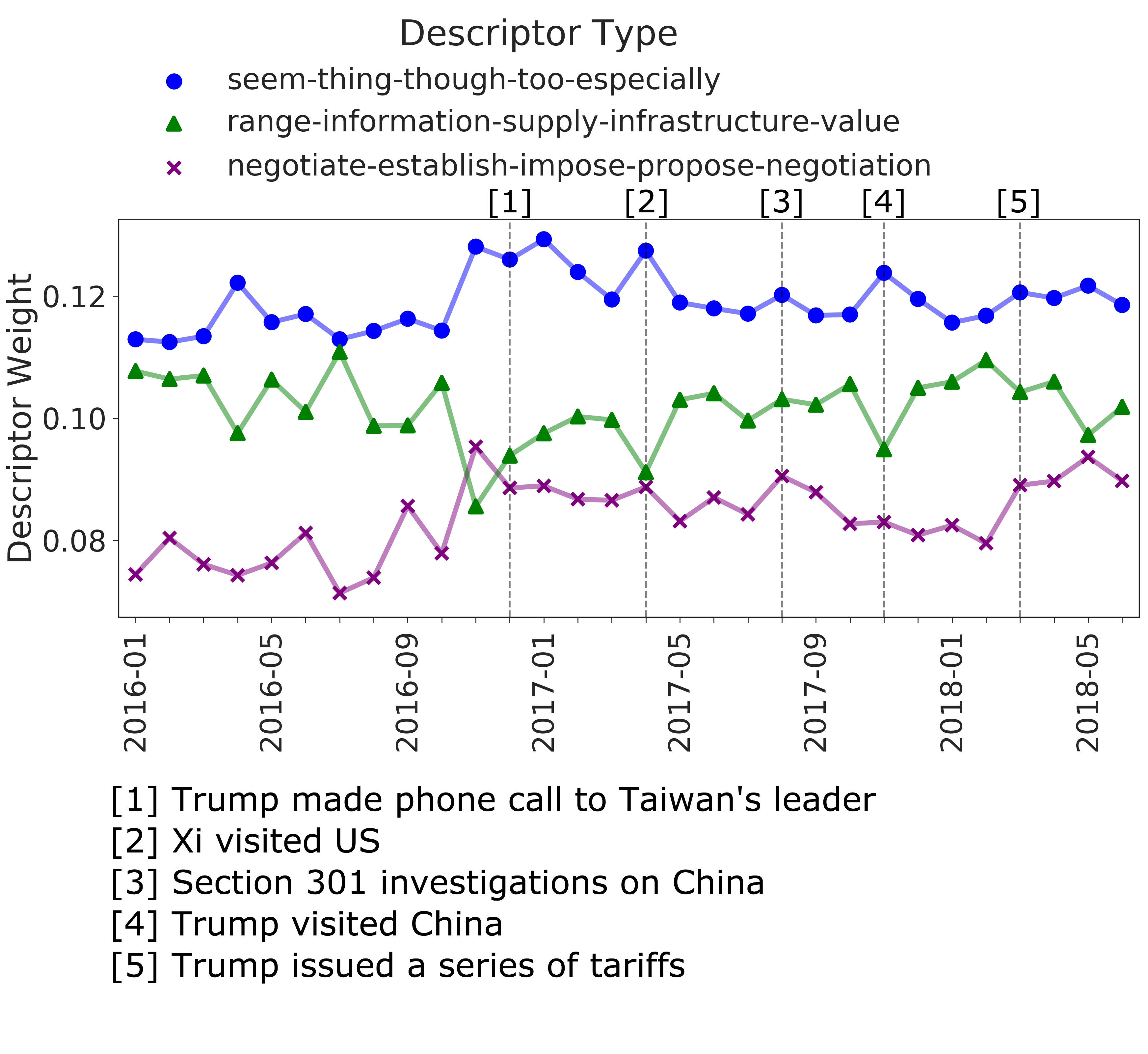}
    \caption{US-China's relation trends by \rmn.}
    \label{fig:trend_mohit}
    \end{subfigure}
    \caption{
    Temporal trends of top three relations between US and China based on \modelname (Figure~\ref{fig:trend_ours}), in comparison with results from \rmn (Figure~\ref{fig:trend_mohit}).
    We highlight key events during this time period under the x-axes.
    Similar visualizations are also used in human evaluations. More figures are provided in the appendix.
    }
    \label{fig:trends}
\end{figure*}

In this section, we compare our model to \rmn. For both models, we fixed the number of descriptors to 30 following \citet{Iyyer:16}. As tracking dynamic %
international relations requires careful analysis, we hosted onsite user studies for quality control and in-person feedback. We first describe the model outputs, and then present both quantitative and qualitative evaluation results.

\subsection{Understanding Model Outputs}
Given a set of time-stamped news articles and a list of nations of interest,
both models provide a set of relation descriptors, where each one \textbf{defines} a type of relation and a temporal trend analysis of these descriptors that shows how the relation \textbf{evolves}.

\para{Relation descriptors.}
Table~\ref{tab:descriptor} shows the top five and bottom five descriptors from \modelname and \rmn sorted by the average weights over all news articles 
related to the most frequently mentioned eight nation pairs.\footnote{We focus on the top eight nation pairs to be consistent with human evaluation.} %
By using predicates to describe relations, our descriptors seem to contain more semantically meaningful words. %
For instance, the top relation in \rmn consists of exclusively non-content words.
Another interesting advantage of our model is that 
the five relations with the lowest weight have much higher weights in \modelname than in \rmn.
This suggests that \rmn tends to generate ``useless'' relations that do not show up in the data, while even bottom relations in \modelname remain useful for describing the data.

\para{Temporal trends.}
We visualize the temporal trends of the most prominent relations between nation pairs. We further provide our annotated key events as the context to interpret these temporal trends.
Figure~\ref{fig:trends} gives an example for US and China. The top three relations based on \modelname are ``denounce'', ``strengthen'', and ``leave'',\footnote{For space reasons, we only include the top word in the relation descriptor.} while the top three based on \rmn are ``seem'', ``range'', and ``negotiate''.
We find our model generally aligns better with the key events: for instance, the ``denounce'' relation peaked around the time that Trump started issuing a series of tariffs based on our model (Figure~\ref{fig:trend_ours}), while there do not exist similar fluctuations in Figure~\ref{fig:trend_mohit} based on \rmn.

\subsection{Quantitative Evaluation} 

\begin{figure}[t]
    \centering
  \includegraphics[width=0.45\textwidth]{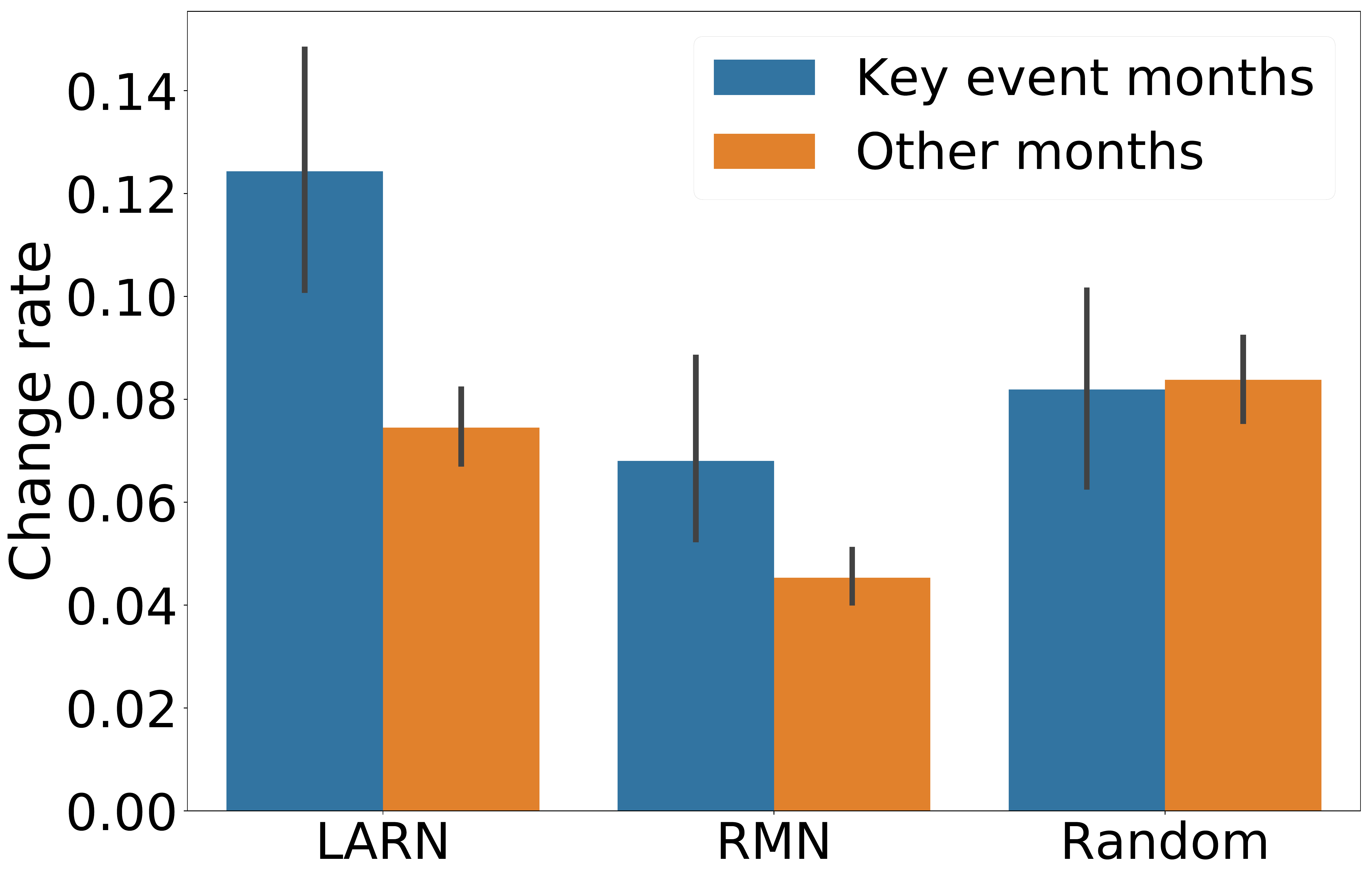}
  \caption{Change rate at key event months vs. other months. Our model demonstrates a clearer relative difference between key event months and other months (66.9\%) than \rmn (50.1\%). For the random baseline, we randomly choose the same number of key events and we observe no differences as expected.
  }\label{fig:cpa}
\end{figure}
We leverage our manually annotated key events to develop a novel automatic 
metric to evaluate how well the temporal trends from our model are aligned with key events.
We define a change rate at each month $\Delta_t$ as the weighted average of changes in relation weights in the top three relations:
$${\Delta_t}  = \sum_{i \in \text{top 3 relations}} w_{t,i} * \frac{|d_{t,i} - d_{t_{prev},i}|}{d_{t_{prev},i}},$$
\noindent where $d_{t,i}$ is the average weight for the $i$-th relation in all the articles published at $t$, $d_{t_{prev}, i}$ is the average weight before $t$ 
in a window of $W$ preceding months ($\frac{1}{W} \sum_{w=1}^{W} d_{(t-w), i}$), and weight $w_{t, i}$ is the normalized weight for top three descriptors ($\frac{d_{t,i}}{\sum_{j \in \text{top 3 relations}} d_{t,j}}$). Our results are robust to the choices of $W$, and we set it to 6 for presentation.

We expect change rates to be greater when significant events happen in international politics. Figure~\ref{fig:cpa} compares the change rate at months where key events occurred with other months for eight nation pairs for which we annotated key events. 
Both models present more abrupt changes when key events occurred.
Unlike \rmn, our model 
does not have temporal dependencies between relation distributions over time,\footnote{Please refer to ~\citet{Iyyer:16}'s Section 3.2.3 for more details on how they incorporated previous time steps.} and thus has a higher discontinuity in general.
However, even in relative terms, our model fluctuated more substantially than \rmn
when key events occurred.
We also did a robustness check with another set of independently annotated key events and the results can be found in the appendix.
This measure captures whether the model can detect the change points, but does not measure whether the model correctly captures the \textbf{semantics} of the key events, i.e., 
did a negative relation increase after hostile events such as war? To this end, we performed human evaluations. 

\subsection{Human Evaluation}
We hosted three human evaluations with participants from different demographics: undergraduate students from 
political science classes,
graduate students from a computer science department (mostly in NLP), and undergraduate students taking a linguistics class. The total number of participants was 29, roughly equally divided among the three groups. The participants were shown outputs from \rmn and \modelname, and asked to choose the output that better aligns with their intuitions. 
Each participant answered about 10 questions and provided justification for their answers to each question, taking roughly 30 minutes to an hour.
Table~\ref{tab:human_eval} summarizes the results of our human evaluations.

\begin{table}[t]
\small
\begin{center}
\begin{tabular}{l|r|r|r}
\toprule
& \multicolumn{2}{c|}{Model} & \multirow{2}{*}{P Value} \\
 & \modelname & \rmn  &   \\
\midrule
Descriptor & 22 & 7 & 8.1e-3 \\
Temporal Trend & 106 & 18 &2.3e-16 \\\midrule
Nation Pair Matching & 38.0 \% & 45.2 \% & 0.33\\%
\bottomrule
\end{tabular} 
\end{center}
\caption{Human evaluation results: The first two rows represents the number of votes, while the last row represents \% of nation pairs matched correctly with its temporal trend. We did a two-tail binomial test for the relation descriptor and temporal trend evaluation and an independent t-test for the nation pair matching evaluation. We show the p-values in the last column.} %
\label{tab:human_eval}
\end{table}

\para{Relation descriptor evaluation.} The participants were shown a list of top five descriptors (as in Table~\ref{tab:descriptor}) from two models, and prompted to select a set which adequately covers possible relationships that can occur between countries.
75.9\% of participants preferred our model.

\para{Temporal trends evaluation.} We showed temporal trends between nation pairs annotated with key events, one from \rmn and the other from \modelname (as in Figure~\ref{fig:trends}). We asked them to evaluate whether the temporal trends accurately reflect the dynamics in nation-to-nation relationships. Each participant evaluated four 
randomly chosen nation pairs. The temporal trend from our model was preferred more frequently (85.5\% of total responses). 

\para{Nation pair matching.} 
We designed a novel task where we showed the participants the other four
temporal trends without annotated key events and asked them to match each trend with the 
corresponding nation pair from the four candidates,
based on their world knowledge about nation-nation relations. 
Each participant did the matching twice, once for \rmn and once for \modelname. The participants found correct temporal trends for 45.2\% of entity pairs for \rmn, and 38.0\% of entity pairs for \modelname, when random pairing would yield 25.0\%. 
The difference between two models here is not statistically significant. Most participants found this task very challenging, as they did not know much about the relationship between certain entity pairs (e.g., a participant said ``As an American, there's no way to know the relation between China and India.'').
Even political science students do not perform better than the other two groups.

\para{Discussion.}
Overall the output from our model is preferred by the participants. 
We found that political science students paid more attention to detail, took a longer time to finish, and were more ambivalent between the performance of the two models. For example, for temporal trends, they preferred our model for 71.4\% of the examples, compared to other groups which preferred our model for 90\% of the examples. They also preferred \rmn's relation descriptors slightly (42.9\% selected our model) and commented that a few concepts from \rmn, like ``infrastructure, supply, and value", are more concrete (e.g., a participant said ``I chose the left one (\rmn), because it is easy to determine and remember the positive of items such as infrastructure, value, and supply. Those have more positive undertones, while it is easy to gauge negative sentiments with `terrorism,' `condemn' and those.''). As \modelname encodes the background context specific to each nation pair separately, our relation descriptors do not contain such ``concrete'' concepts. In the next section, we will discuss \textbf{contexts} for relations, where these concepts appear in \modelname. %

\section{Further Exploration}
\begin{figure}[t]
  \centering
  \includegraphics[width=0.48\textwidth]{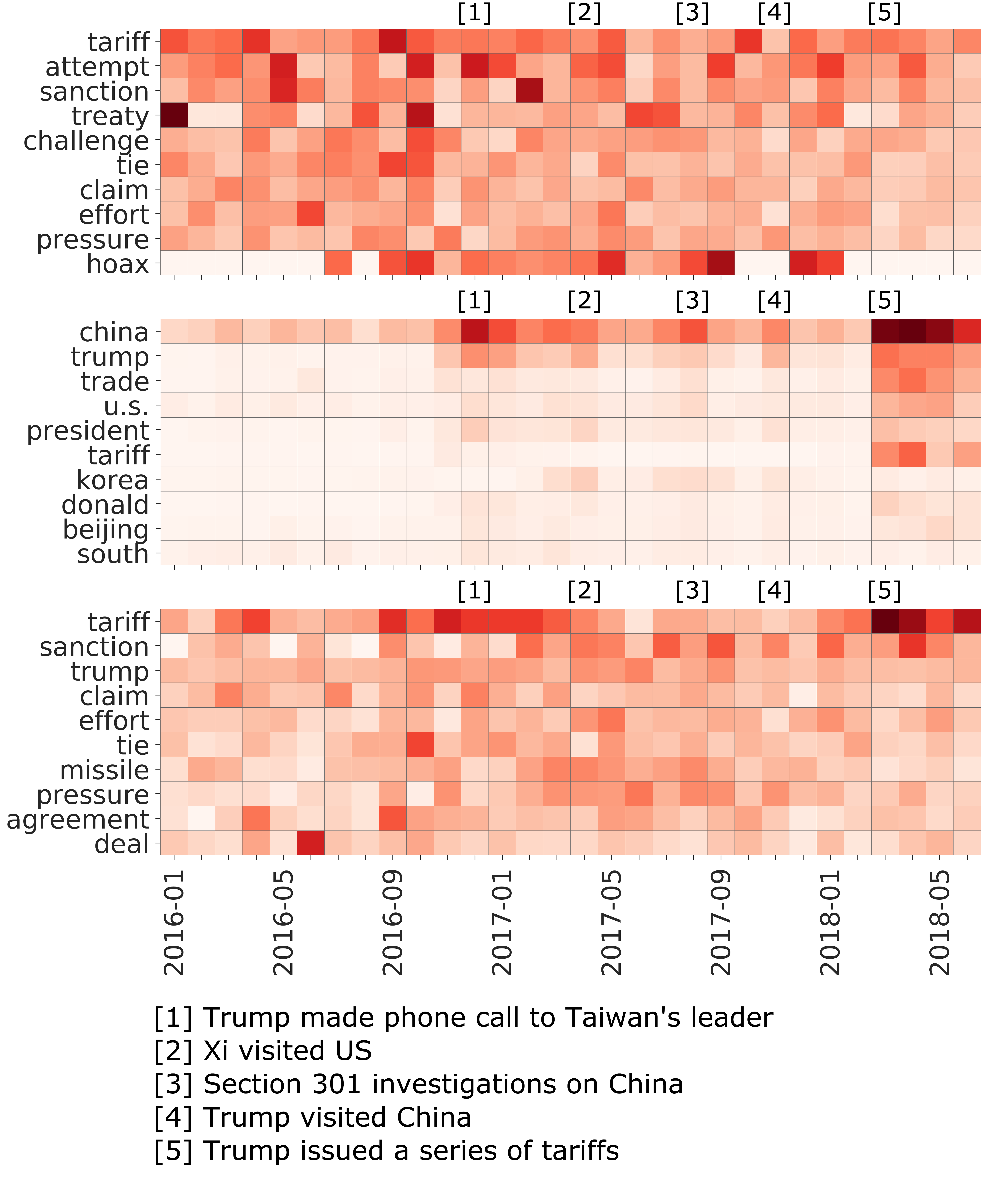}
  \caption{Top contextual words for US-China's ``denounce" relation derived from three approaches. \textit{Top}: Showing word's avg. attention score in all ``denounce'' articles each month. \textit{Middle}: Showing word's frequency in all ``denounce'' articles each month. \textit{Bottom}: Showing word's average attention score multiplied by $\log(\text{appearance})$ in each month. All figures are normalized by the global maximum score in the figure.
  }
  \label{fig:aw}
\end{figure}

We present additional qualitative results to showcase the applications of our model.
First, we examine the context (nouns) associated with a relation between two nations based on an attention-based mechanism, which \rmn does not handle.
Second, we perform an in-depth analysis to show how our model can reveal regional differences in news coverage on the same topic.

\subsection{Context for Relations}%
To help users better understand the inferred relations, we offer specific contexts that relate to the inferred relations based on the attention-based mechanism introduced in \S\ref{sec:model}.
For each relation, we find articles 
that place the most
weight in that relation, and rank the nouns and proper nouns in those articles by their 
average attention score (i.e., $\alpha_{n^a_i}$ in \S\ref{subsec:model_desc}). %

The top part of Figure~\ref{fig:aw} shows the top 10 words for the ``denounce'' relation in \figref{fig:trend_ours}, the temporal trend for US-China relations.
Since \rmn does not support such mechanism, we show the most frequent nouns that occurred in the documents that mention both entities as a baseline (middle part).\footnote{We had a comparison between the top figure and the middle figure in the human evaluation, but we found an error in visualization and thus focus on qualitative comparisons.}
We find that the attended nouns from our model are more informative than frequent nouns:
``tariff'' is the most attended word; words such as ``sanction'', ``treaty'', and ``pressure'' also show up, while the frequency baseline centers around %
words like ``China'' and ``president''.
As we noticed that the frequency baseline can capture alignment with key events, we incorporate attention score and frequency in the bottom part of Figure~\ref{fig:aw}.
This augmented version captures informative words (e.g., ``tariff'', ``sanction'', and ``missile'')
\footnote{``missile'' points us to another event related to US deploying missile defense system in South Korea, which also impacts US-China relations.}
and closely aligns with the key events.%

\subsection{Regional Differences in News Coverage}

To further demonstrate the utility of our model, we explore regional differences in news coverage,
as ``it is possible to build real knowledge by comparing perspectives from different social contexts."\footnote{https://www.publicbooks.org/why-an-age-of-machine-learning-needs-the-humanities/}
This also relates to the longstanding literature on framing in news coverage, i.e., ``selecting some aspects of a perceived reality and make them more salient to promote problem definition/interpretation'' \cite{entman1993framing,chong2007framing}.

We picked two countries, Singapore and US, to study US-China relations.\footnote{China is not an English speaking country and is thus not in the NOW corpus. Singapore contained the most articles containing both US and China.} Using 
country source of media outlets in the NOW corpus, we found 10K articles from Singaporean media and 5.7K articles from US media on US and China.

Table~\ref{tab:diff} shows the top five relations sorted by their absolute weight differences between US media outlets and Singaporean media outlets.
Singaporean media more frequently use ``positive'' descriptors such as ``strengthen'' and ``purchase'', whereas US media report negative relations such as ``denounce'' and ``criticize'' more frequently. %
Table~\ref{tab:example} shows two example sentences from articles with the most weight in the ``denounce'' relation.
Even though two media sources are focusing on events leading to the same type of relation, a reader who mainly consume news articles in Singapore would get a clearly different impression of US-China relations from those who read US news.

\begin{table}[t]
\small
\begin{center}
\renewcommand{\arraystretch}{1.1}
\begin{tabular}{p{0.25\textwidth}rr}
\toprule
 Relation descriptor & US & Singapore\\
\midrule
{strengthen, enhance, improve, develop, boost} & 4.62\% & \textbf{5.42\%}\\
{purchase, sell, pay, buy, cost} & 3.18\% & \textbf{3.86}\%\\
{denounce, undermine, condemn, punish, oppose} & \textbf{6.21}\% & 5.66\%\\
{wag, criticise, accuse, tell, criticize} & \textbf{3.81}\% & 3.29\%\\
{flee, destroy, invade, unmake, expel} & \textbf{3.70}\% & 3.32\%\\
\bottomrule
\end{tabular} 
\end{center}
\caption{
Top five relations between US and China sorted by absolute weight differences between US media outlets and Singaporean media outlets.
}
\label{tab:diff}
\end{table}

\begin{table}
\begin{center}
\begin{tabular}{p{0.45\textwidth}}
\toprule
US media: ``President Donald Trump is preparing to \textcolor{red}{\em impose} a \textcolor{cyan}{\em package} of \$60 billion in annual \textcolor{cyan}{\em tariffs} against Chinese products, following through on a longtime \textcolor{cyan}{\em threat} that he says will \textcolor{red}{\em punish} China for intellectual property theft and \textcolor{red}{\em create} more American jobs.''\\
\midrule
Singaporean media: 
It does not look like just a \textcolor{cyan}{\em trade} \textcolor{cyan}{\em war}, but rather the US is trying to \textcolor{red}{\em bully} China and the rest of the world in order to make China \textcolor{red}{\em concede} economic resources and development \textcolor{cyan}{\em opportunities} to the US and make the US forever big and strong.\\
\bottomrule
\end{tabular} 
\end{center}
\caption{
Example sentences in top-scoring articles on US-China's ``denounce'' relation in March 2018.
We italicize key predicates (red color) and three most attended nouns (cyan color).
}
\label{tab:example}
\end{table}

\section{Related Work}
Prior work~\cite{Chambers2015IdentifyingPS,Choi2016DocumentlevelSI,Rashkin2017MultilingualCF} studied entity-entity relations in terms of positive and negative sentiments between them. Similarly, literature on relation extraction~\cite{riedel2010modeling,Gardner2017OpenVocabularySP,Elson2010ExtractingSN,Srivastava2016InferringIR} focused on pre-defined relations between a pair of entities in the database schema. 
In comparison, our work discovers descriptors for relations between entity pairs instead of finding entity pairs matching pre-defined relation schema.

Topic modeling has been an important method to grasp important concepts from a large collection of documents in an unsupervised fashion~\cite{blei2003latent,Das2015GaussianLF,Chang2009ConnectionsBT,schein2015bayesian}.
Similar to our work, \citet{o2013learning} incorporates linguistic insights with topic models to identify event classes and detect conflicts.
Our work additionally models the context of relations through nouns and focuses on exploring the potential of neural models.
Most relevant to our work is \citet{Iyyer:16},
which suggests \rmn better capture dynamic relationships in literature than hidden Markov model~\cite{Gruber2007HiddenTM} and LDA~\cite{blei2003latent}. %
Recent work extended and applied \rmn to other settings such as studying user roles in online communities \cite{Wang2016LearningLD,Frermann2017InducingSM}.
Notably, \citet{Chaturvedi2017UnsupervisedLO} suggests HMM with shallow linguistic features (i.e., frame net parses) and global constraints can outperform \rmn for modeling relations in literature.
In this work, we incorporate linguistic insights with \rmn and apply it to news domain. 

Last but not least, researchers have studied the dynamics of media coverage from a wide range of perspectives,
ranging from framing~\cite{Card2015TheMF,field2018framing},
to relationship between ideas~\cite{Tan2017FriendshipsRA},
to quotes of politicians~\cite{niculae2015quotus,tan2018you,leskovec2009meme}.
There is also significant effort for building event databases in political science~\cite{leetaru2013gdelt}, and assisting journalists with tools~\cite{Handler2017RookieAU}, and dating historical text~\cite{Niculae2014TemporalTR}.

\section{Conclusion}
We investigate the promise of unsupervised neural models for automatically inferring relations between nations.
We find that incorporating shallow linguistic information is a simple yet effective strategy for deriving robust and interpretable relations.
We develop a novel quantitative evaluation metric for understanding international relations and in-depth human evaluation also confirms the effectiveness of our model.
We further show that our models can provide the background of relations 
using attention score and reveal regional differences for future studies on media framing.

Meanwhile, our work suggests important future directions for using NLP technologies to support individuals in navigating a large collection of news articles.
Our participants often find it challenging to infer information simply from temporal dynamics of our inferred relations based on natural language descriptors.
It is thus important to incorporate human cognitive preferences in developing such models and provide narratives beyond words such as key events.
Furthermore, different populations pay attention to different parts of information. 
We need to understand the diversity when developing NLP technologies for end users and provide helpful personalized hints to lower the barrier of benefiting from model outputs.

\section*{Acknowledgments}
We thank the anonymous reviewers, Dallas Card, and Mohit Iyyer for helpful feedback and discussion.
We thank all the participants in our human evaluation from LING 3813 at Georgia Tech by Lelia Glass, 
students at X-lab at University of Washington, and undergraduate students at University of Colorado Boulder.
We also thank Andy Baker and Sven Steinmo for recruiting students in their classes.
Choi is supported by a Facebook Fellowship.

%
%
%
%
%


\begin{thebibliography}{38}
\expandafter\ifx\csname natexlab\endcsname\relax\def\natexlab#1{#1}\fi

\bibitem[{Bahdanau et~al.(2015)Bahdanau, Cho, and Bengio}]{Bahdanau}
Dzmitry Bahdanau, Kyunghyun Cho, and Yoshua Bengio. 2015.
\newblock Neural machine translation by jointly learning to align and
  translate.
\newblock In \emph{ICLR}.

\bibitem[{Baylis et~al.(2017)Baylis, Smith, and
  Owens}]{baylis2017globalization}
John Baylis, Steve Smith, and Patricia Owens. 2017.
\newblock \emph{The globalization of world politics: an introduction to
  international relations}.
\newblock Oxford University Press.

\bibitem[{Blei et~al.(2003)Blei, Ng, and Jordan}]{blei2003latent}
David~M Blei, Andrew~Y Ng, and Michael~I Jordan. 2003.
\newblock Latent dirichlet allocation.
\newblock \emph{Journal of machine Learning research}, 3(Jan):993--1022.

\bibitem[{Card et~al.(2015)Card, Boydstun, Gross, Resnik, and
  Smith}]{Card2015TheMF}
Dallas Card, Amber~E. Boydstun, Justin~H. Gross, Philip Resnik, and Noah~A.
  Smith. 2015.
\newblock The media frames corpus: Annotations of frames across issues.
\newblock In \emph{ACL}.

\bibitem[{Chambers et~al.(2015)Chambers, Bowen, Genco, Tian, Young, Harihara,
  and Yang}]{Chambers2015IdentifyingPS}
Nathanael Chambers, Victor Bowen, Ethan Genco, Xisen Tian, Eric Young, Ganesh
  Harihara, and Eugene Yang. 2015.
\newblock Identifying political sentiment between nation states with social
  media.
\newblock In \emph{EMNLP}.

\bibitem[{Chaney et~al.(2016)Chaney, Wallach, Connelly, and
  Blei}]{Chaney2016DetectingAC}
Allison June-Barlow Chaney, Hanna~M. Wallach, Matthew Connelly, and David~M.
  Blei. 2016.
\newblock Detecting and characterizing events.
\newblock In \emph{EMNLP}.

\bibitem[{Chang et~al.(2009)Chang, Boyd-Graber, and
  Blei}]{Chang2009ConnectionsBT}
Jonathan Chang, Jordan~L. Boyd-Graber, and David~M. Blei. 2009.
\newblock Connections between the lines: augmenting social networks with text.
\newblock In \emph{KDD}.

\bibitem[{Chaturvedi et~al.(2017)Chaturvedi, Iyyer, and
  Daum{\'e}}]{Chaturvedi2017UnsupervisedLO}
Snigdha Chaturvedi, Mohit Iyyer, and Hal Daum{\'e}. 2017.
\newblock Unsupervised learning of evolving relationships between literary
  characters.
\newblock In \emph{AAAI}.

\bibitem[{Choi et~al.(2016)Choi, Rashkin, Zettlemoyer, and
  Choi}]{Choi2016DocumentlevelSI}
Eunsol Choi, Hannah Rashkin, Luke~S. Zettlemoyer, and Yejin Choi. 2016.
\newblock Document-level sentiment inference with social, faction, and
  discourse context.
\newblock In \emph{ACL}.

\bibitem[{Chong and Druckman(2007)}]{chong2007framing}
Dennis Chong and James~N Druckman. 2007.
\newblock Framing theory.
\newblock \emph{Annual Review of Political Science}, 10:103--126.

\bibitem[{Conneau et~al.(2017)Conneau, Kiela, Schwenk, Barrault, and
  Bordes}]{Conneau2017SupervisedLO}
Alexis Conneau, Douwe Kiela, Holger Schwenk, Lo{\"i}c Barrault, and Antoine
  Bordes. 2017.
\newblock Supervised learning of universal sentence representations from
  natural language inference data.
\newblock In \emph{EMNLP}.

\bibitem[{Das et~al.(2015)Das, Zaheer, and Dyer}]{Das2015GaussianLF}
Rajarshi Das, Manzil Zaheer, and Chris Dyer. 2015.
\newblock Gaussian lda for topic models with word embeddings.
\newblock In \emph{ACL}.

\bibitem[{Doddington et~al.(2004)Doddington, Mitchell, Przybocki, Ramshaw,
  Strassel, and Weischedel}]{doddington2004automatic}
George~R Doddington, Alexis Mitchell, Mark~A Przybocki, Lance~A Ramshaw,
  Stephanie Strassel, and Ralph~M Weischedel. 2004.
\newblock The automatic content extraction (ace) program-tasks, data, and
  evaluation.
\newblock In \emph{LREC}.

\bibitem[{Elson et~al.(2010)Elson, Dames, and McKeown}]{Elson2010ExtractingSN}
David~K. Elson, Nicholas Dames, and Kathleen McKeown. 2010.
\newblock Extracting social networks from literary fiction.
\newblock In \emph{ACL}.

\bibitem[{Entman(1993)}]{entman1993framing}
Robert~M Entman. 1993.
\newblock Framing: Toward clarification of a fractured paradigm.
\newblock \emph{Journal of communication}, 43(4):51--58.

\bibitem[{Fellbaum(1998)}]{wordnet}
Christiane Fellbaum. 1998.
\newblock \emph{WordNet: An Electronic Lexical Database}.
\newblock Bradford Books.

\bibitem[{Field et~al.(2018)Field, Kliger, Wintner, Pan, Jurafsky, and
  Tsvetkov}]{field2018framing}
Anjalie Field, Doron Kliger, Shuly Wintner, Jennifer Pan, Dan Jurafsky, and
  Yulia Tsvetkov. 2018.
\newblock Framing and agenda-setting in russian news: a computational analysis
  of intricate political strategies.
\newblock In \emph{EMNLP}.

\bibitem[{Frermann and Szarvas(2017)}]{Frermann2017InducingSM}
Lea Frermann and Gy{\"o}rgy Szarvas. 2017.
\newblock Inducing semantic micro-clusters from deep multi-view representations
  of novels.
\newblock In \emph{EMNLP}.

\bibitem[{Gardner and Krishnamurthy(2017)}]{Gardner2017OpenVocabularySP}
Matt Gardner and Jayant Krishnamurthy. 2017.
\newblock Open-vocabulary semantic parsing with both distributional statistics
  and formal knowledge.
\newblock In \emph{AAAI}.

\bibitem[{Glorot and Bengio(2010)}]{pmlr-v9-glorot10a}
Xavier Glorot and Yoshua Bengio. 2010.
\newblock Understanding the difficulty of training deep feedforward neural
  networks.
\newblock In \emph{AISTATS}.

\bibitem[{Gruber et~al.(2007)Gruber, Weiss, and Rosen-Zvi}]{Gruber2007HiddenTM}
Amit Gruber, Yair Weiss, and Michal Rosen-Zvi. 2007.
\newblock Hidden topic markov models.
\newblock In \emph{AISTATS}.

\bibitem[{Handler and O'Connor(2017)}]{Handler2017RookieAU}
Abram Handler and Brendan~T. O'Connor. 2017.
\newblock Rookie: A unique approach for exploring news archives.
\newblock In \emph{Workshop on Data Science + Journalism at KDD}.

\bibitem[{Honnibal and Montani(2017)}]{spacy2}
Matthew Honnibal and Ines Montani. 2017.
\newblock spacy 2: Natural language understanding with bloom embeddings,
  convolutional neural networks and incremental parsing.

\bibitem[{Iyyer et~al.(2016)Iyyer, Guha, Chaturvedi, Boyd-Graber, and
  {Daum\'{e} III}}]{Iyyer:16}
Mohit Iyyer, Anupam Guha, Snigdha Chaturvedi, Jordan Boyd-Graber, and Hal
  {Daum\'{e} III}. 2016.
\newblock Feuding families and former friends: Unsupervised learning for
  dynamic fictional relationships.
\newblock In \emph{NAACL}.

\bibitem[{Leetaru and Schrodt(2013)}]{leetaru2013gdelt}
Kalev Leetaru and Philip~A Schrodt. 2013.
\newblock Gdelt: Global data on events, location, and tone, 1979--2012.
\newblock In \emph{ISA annual convention}.

\bibitem[{Leskovec et~al.(2009)Leskovec, Backstrom, and
  Kleinberg}]{leskovec2009meme}
Jure Leskovec, Lars Backstrom, and Jon Kleinberg. 2009.
\newblock Meme-tracking and the dynamics of the news cycle.
\newblock In \emph{KDD}.

\bibitem[{Mintz et~al.(2009)Mintz, Bills, Snow, and
  Jurafsky}]{mintz2009distant}
Mike Mintz, Steven Bills, Rion Snow, and Dan Jurafsky. 2009.
\newblock Distant supervision for relation extraction without labeled data.
\newblock In \emph{ACL}.

\bibitem[{Niculae et~al.(2015)Niculae, Suen, Zhang, Danescu-Niculescu-Mizil,
  and Leskovec}]{niculae2015quotus}
Vlad Niculae, Caroline Suen, Justine Zhang, Cristian Danescu-Niculescu-Mizil,
  and Jure Leskovec. 2015.
\newblock Quotus: The structure of political media coverage as revealed by
  quoting patterns.
\newblock In \emph{WWW}.

\bibitem[{Niculae et~al.(2014)Niculae, Zampieri, Dinu, and
  Ciobanu}]{Niculae2014TemporalTR}
Vlad Niculae, Marcos Zampieri, Liviu~P. Dinu, and Alina~Maria Ciobanu. 2014.
\newblock Temporal text ranking and automatic dating of texts.
\newblock In \emph{EACL}.

\bibitem[{O'Connor et~al.(2013)O'Connor, Stewart, and Smith}]{o2013learning}
Brendan O'Connor, Brandon~M Stewart, and Noah~A Smith. 2013.
\newblock Learning to extract international relations from political context.
\newblock In \emph{ACL}.

\bibitem[{Pennington et~al.(2014)Pennington, Socher, and
  Manning}]{pennington2014glove}
Jeffrey Pennington, Richard Socher, and Christopher~D. Manning. 2014.
\newblock Glove: Global vectors for word representation.
\newblock In \emph{EMNLP}.

\bibitem[{Rashkin et~al.(2017)Rashkin, Bell, Choi, and
  Volkova}]{Rashkin2017MultilingualCF}
Hannah Rashkin, Eric Bell, Yejin Choi, and Svitlana Volkova. 2017.
\newblock Multilingual connotation frames: A case study on social media for
  targeted sentiment analysis and forecast.
\newblock In \emph{ACL}.

\bibitem[{Riedel et~al.(2010)Riedel, Yao, and McCallum}]{riedel2010modeling}
Sebastian Riedel, Limin Yao, and Andrew McCallum. 2010.
\newblock Modeling relations and their mentions without labeled text.
\newblock In \emph{ECML}.

\bibitem[{Schein et~al.(2015)Schein, Paisley, Blei, and
  Wallach}]{schein2015bayesian}
Aaron Schein, John Paisley, David~M Blei, and Hanna Wallach. 2015.
\newblock Bayesian poisson tensor factorization for inferring multilateral
  relations from sparse dyadic event counts.
\newblock In \emph{KDD}.

\bibitem[{Srivastava et~al.(2016)Srivastava, Chaturvedi, and
  Mitchell}]{Srivastava2016InferringIR}
Shashank Srivastava, Snigdha Chaturvedi, and Tom~M. Mitchell. 2016.
\newblock Inferring interpersonal relations in narrative summaries.
\newblock In \emph{AAAI}.

\bibitem[{Tan et~al.(2017)Tan, Card, and Smith}]{Tan2017FriendshipsRA}
Chenhao Tan, Dallas Card, and Noah~A. Smith. 2017.
\newblock Friendships, rivalries, and trysts: Characterizing relations between
  ideas in texts.
\newblock In \emph{ACL}.

\bibitem[{Tan et~al.(2018)Tan, Peng, and Smith}]{tan2018you}
Chenhao Tan, Hao Peng, and Noah~A Smith. 2018.
\newblock " you are no jack kennedy": On media selection of highlights from
  presidential debates.
\newblock In \emph{WWW}.

\bibitem[{Wang et~al.(2016)Wang, Hamilton, and Leskovec}]{Wang2016LearningLD}
Alex Wang, William~L. Hamilton, and Jure Leskovec. 2016.
\newblock Learning linguistic descriptors of user roles in online communities.
\newblock In \emph{NLP+CSS@EMNLP}.

\end{thebibliography}

\bibliographystyle{acl_natbib}

\clearpage
\appendix

\newcommand{\hbAppendixPrefix}{A}
\renewcommand{\thefigure}{\hbAppendixPrefix\arabic{figure}}
\setcounter{figure}{0}
\renewcommand{\thetable}{\hbAppendixPrefix\arabic{table}} 
\setcounter{table}{0}

\section{Appendix}

\subsection*{Data: Aliases and Key Events}
We use a dictionary of nation aliases to find relevant sentences in the corpus that mentioned at least one nation pair.\footnote{We did not use coreference resolution to avoid noisy linking.} Table~\ref{tab:aliases} shows the full alias dictionary we used. Table~\ref{tab:events} shows the complete list of key events used in this paper.

\begin{table}[h]
\small
\begin{center}
\renewcommand{\arraystretch}{1.1}
\begin{tabular}{lr}
\toprule
Nation & Aliases\\
\midrule
U.S. & {U.S., US, USA, Trump, Obama}\\
China. & {China, Chinese, Xi}\\
Syria & {Syria, Syrian, Assad}\\
France & {France, French, Macron, Hollande}\\
Germany. & {Germany, German, Merkel}\\
Canada & {Canada, Canadian, Trudeau}\\
Russia & {Russia, Russian, Putin}\\
U.K. & {U.K., UK, British, Britain, Cameron}\\
India & {India, Indian, Modi}\\
Japan & {Japan, Japanese, Abe}\\
Iran & {Iran, Iranian, Khamenei, Rouhani}\\
Israel & {Israel, Israeli, Netanyahu}\\
\bottomrule
\end{tabular} 
\end{center}
\caption{
Aliases used for detecting nation mentions.
}
\label{tab:aliases}
\end{table}

\subsection*{Preprocessing}

We use spaCy to preprocess all sentences~\cite{spacy2}.
We then use a rule based extractor\footnote{\url{https://github.com/NSchrading/intro-spacy-nlp/blob/master/subject_object_extraction.py}} to get verbal predicates with detectable subjects and objects. When verbal predicate has a negation, we take its antonym in WordNet~\cite{wordnet} if it exists, ignored otherwise. We also extract nouns and proper nouns in sentences using spaCy's part-of-speech tagging.

We follow the preprocessing steps in \citet{Iyyer:16} for \rmn. The original \rmn, which used a fiction dataset, removes the 500 most frequently occurring words and words that occur in fewer than 100 books. We also remove the 500 most frequent words for the input to \rmn. However, since our news dataset doesn't have the notion of ``books'', we remove 5000 least common words (out of a ~200K vocabulary) for \rmn.
Note that this is not done for our model \modelname.

\begin{table}[h]
\small
\begin{center}
\renewcommand{\arraystretch}{1.1}
\begin{tabular}{lr}
\toprule
Hyperparameter & Value\\
\midrule
$v_{label}$, $v_{p}$, $v_n$ dimension & 300\\
$p_{drop}$ & 0.5\\
$v_{e}$ dimension & 50\\
$h_{n}$ (\modelname) dimension & 300 + 30\\
$v_{\text{final}}$ dimension & 300\\
recurrent enforcement $\alpha$ (\rmn) & 0.5\\
number of relations $K$& 30\\
number of training epochs & 15\\
learning rate & 1e-3\\
orthogonal penalty $\lambda$ & 1e-1\\
number of negative samples & 15\\
batch size & 256\\
\bottomrule
\end{tabular} 
\end{center}
\caption{
Hyperparameters.
}
\label{tab:hyperparameters}
\end{table}

\subsection*{Implementation Details}
We show all hyperparameters used in our study in Table~\ref{tab:hyperparameters}. We generally use the same hyperparameter values as \rmn, except a few model-specific hyperparameters in Table~\ref{tab:hyperparameters}. We also use Xavier initialization~\cite{pmlr-v9-glorot10a} for all trainable layers as in the original \rmn.

\subsection*{Relation Descriptors Postprocessing}
In addition to finding the nearest neighboring words to the relation embeddings, the orginal \rmn also requires a manual filtering process to achieve an interpretable definition of the relation descriptor. To prevent introducing bias in this manual filtering step, we define each relation descriptor directly as its five nearest neighboring words in the input vocabulary. However, since the full vocabulary would contain many uncommon words which could hinder the interpretation of relation descriptors, we limit both models 
to choose descriptor words from the most frequently occurring 500 words in their own processed input vocabulary (i.e., verb predicates only for \modelname and all words for \rmn).

In Table~\ref{tab:unlimited_descriptor}, we present the descriptor set by \rmn and \modelname without the most common 500 words constraint on descriptor word selection. We find that the descriptor set of our model is robust to this change and does not suffer from a lowered interpretability as much as \rmn.

\begin{table*}[h]
\small
\begin{center}
\begin{tabular}{p{0.37\textwidth}r|p{0.37\textwidth}r}
\toprule
\modelname (Linguistically aware relationship network) & Weight & \rmn (Relationship modeling network) & Weight  \\
\midrule
{denounce, undermine, condemn, decry, legitimize} & 5.46\%&
{always, something, seem, sense, thought} & 11.79\%  \\ {leave,tell,forget,ask,know} & 4.93\%&
{components, utilization, integrated, optimized, component} & 9.42\%\\
{differ, indicate, extent, affect, imply} & 4.64\%&
{morroco, theresa, miriam, emmanuel, charlene} & 8.24\% \\
{strengthen, enhance, improve, develop, maximize} & 4.56\%&
{mandate, authorize, assurances, rescind, ratify} & 7.92\%\\
{hit, fell, slump, edge, cut} & 4.05\%&
{provocation, bloodshed, reprisals, incursion, hostilities} & 7.54\% \\
\midrule
{buy, cheap, manufacture, use, shop} & 2.59\% & {una, esa, harridan, chicas, hija} & 1.69e-09 \\
{win, clinch, defeat, championship, champion} & 2.59\% & {gest, epoque, rol, chiat, tripper} & 1.62e-09 \\
{receive, bestow, redeem, entitle, outpour} & 2.58\% & {pharmacologically, offred, condemnable, chapelle, benassi} & 5.39e-10 \\
{offer, provide, deliver, ideal, cater} & 2.46\% & {ricoh, powe, mw, sy, hamill} & 2.94e-10 \\
{launch, relaunch, announce, debut, unveil} & 2.36\% & 
{tunstall, booksellers, seiko, pring, reflation} & 2.21e-10 \\
\bottomrule
\end{tabular} 
\end{center} \vspace{-1pt}
\caption{Relation descriptors of the most/least frequent five relations by \rmn and \modelname and their average weights in news articles (without constraint on most common words). %
}  
\label{tab:unlimited_descriptor}
\end{table*}

\subsection*{Additional Temporal Relation Trends: US-Russia, US-India, US-Syria}
\begin{figure*}[t]
    \begin{subfigure}[h]{0.47\textwidth}
    \includegraphics[width=\textwidth]{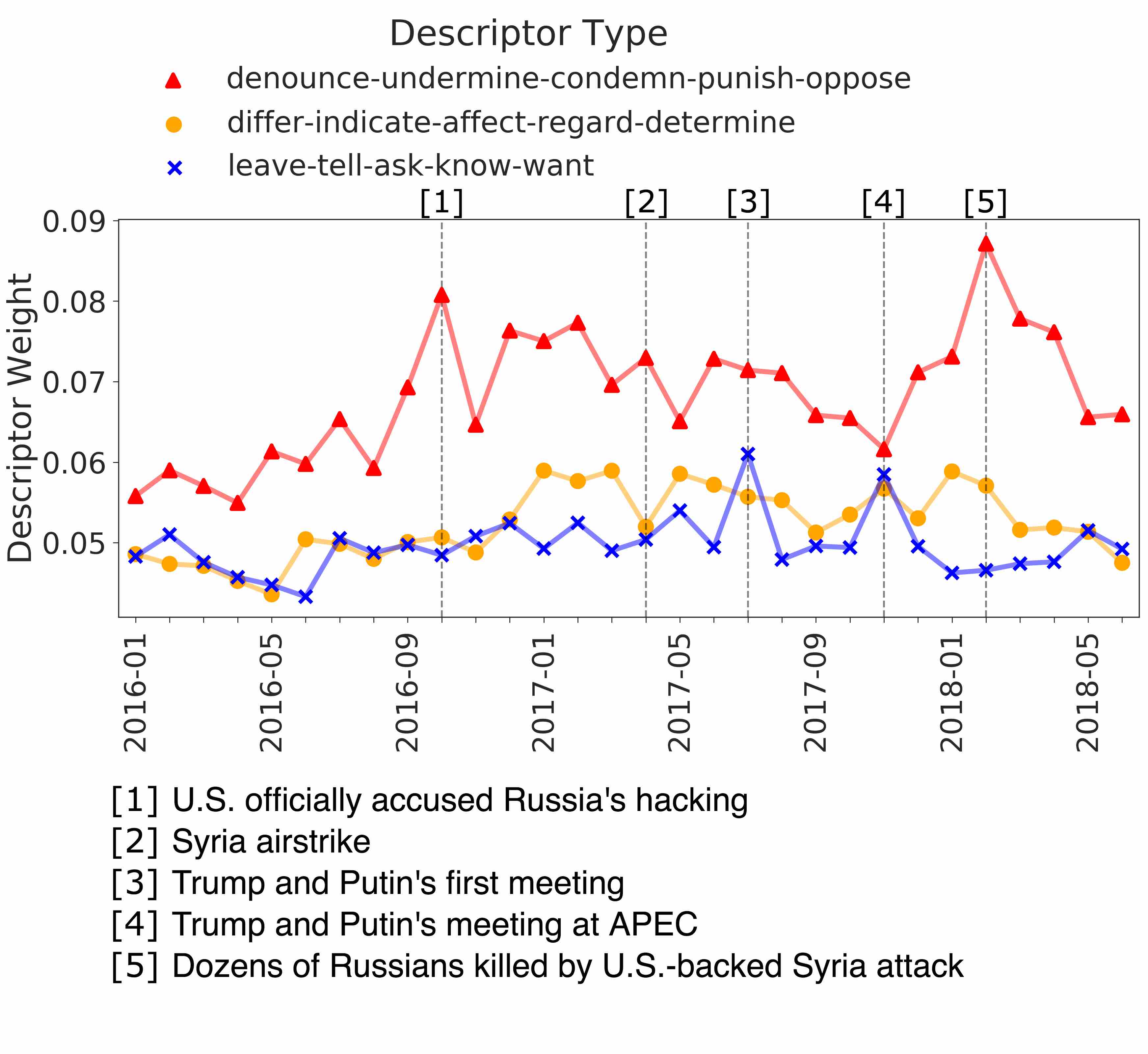}
    \caption{US-Russia's relation trends by \modelname.}
    \label{fig:UsRu_trend_ours}
    \end{subfigure}
    \hfill
    \begin{subfigure}[h]{0.47\textwidth}
    \includegraphics[width=\textwidth]{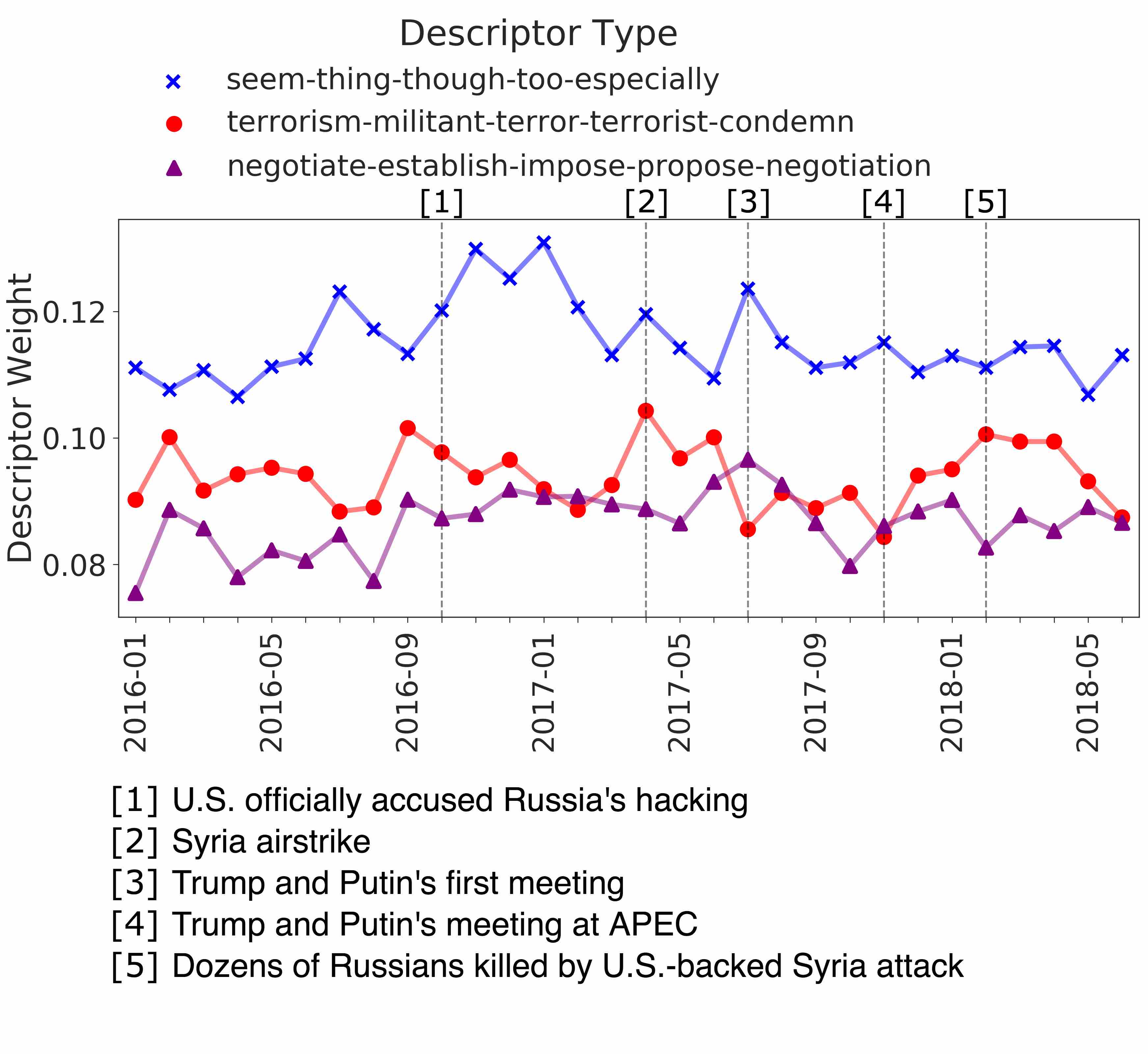}
    \caption{US-Russia's relation trends by \rmn.}
    \label{fig:UsRu_trend_mohit}
    \end{subfigure}
    \caption{
    Temporal trends of top three relations between U.S. and Russia based on \modelname (Figure~\ref{fig:UsRu_trend_ours}), in comparison with results from \rmn (Figure~\ref{fig:UsRu_trend_mohit}).
    }
    \label{fig:UsRu_trends}
\end{figure*}

\begin{figure*}[t]
    \begin{subfigure}[h]{0.47\textwidth}
    \includegraphics[width=\textwidth]{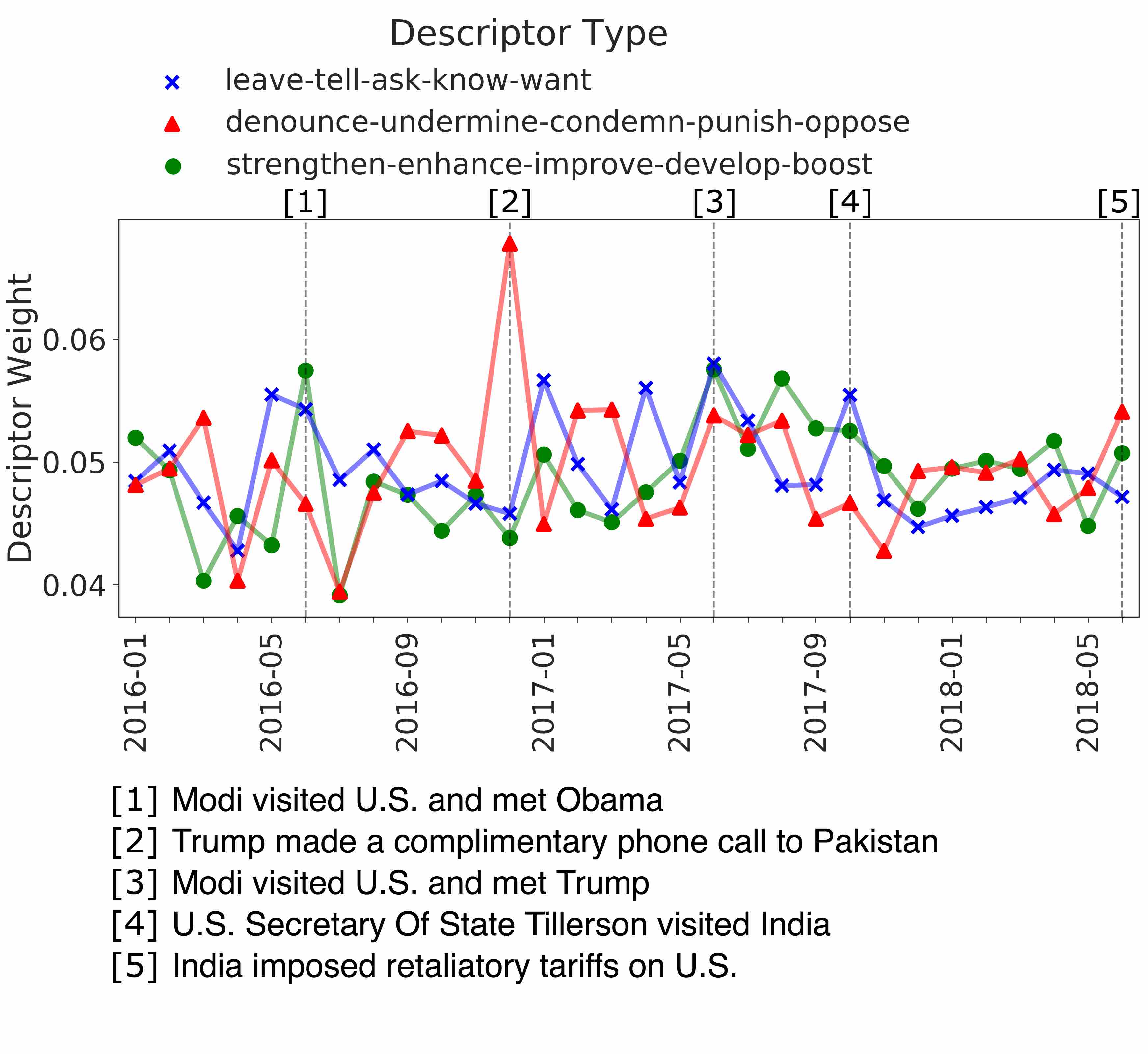}
    \caption{US-India's relation trends by \modelname.}
    \label{fig:UsIn_trend_ours}
    \end{subfigure}
    \hfill
    \begin{subfigure}[h]{0.47\textwidth}
    \includegraphics[width=\textwidth]{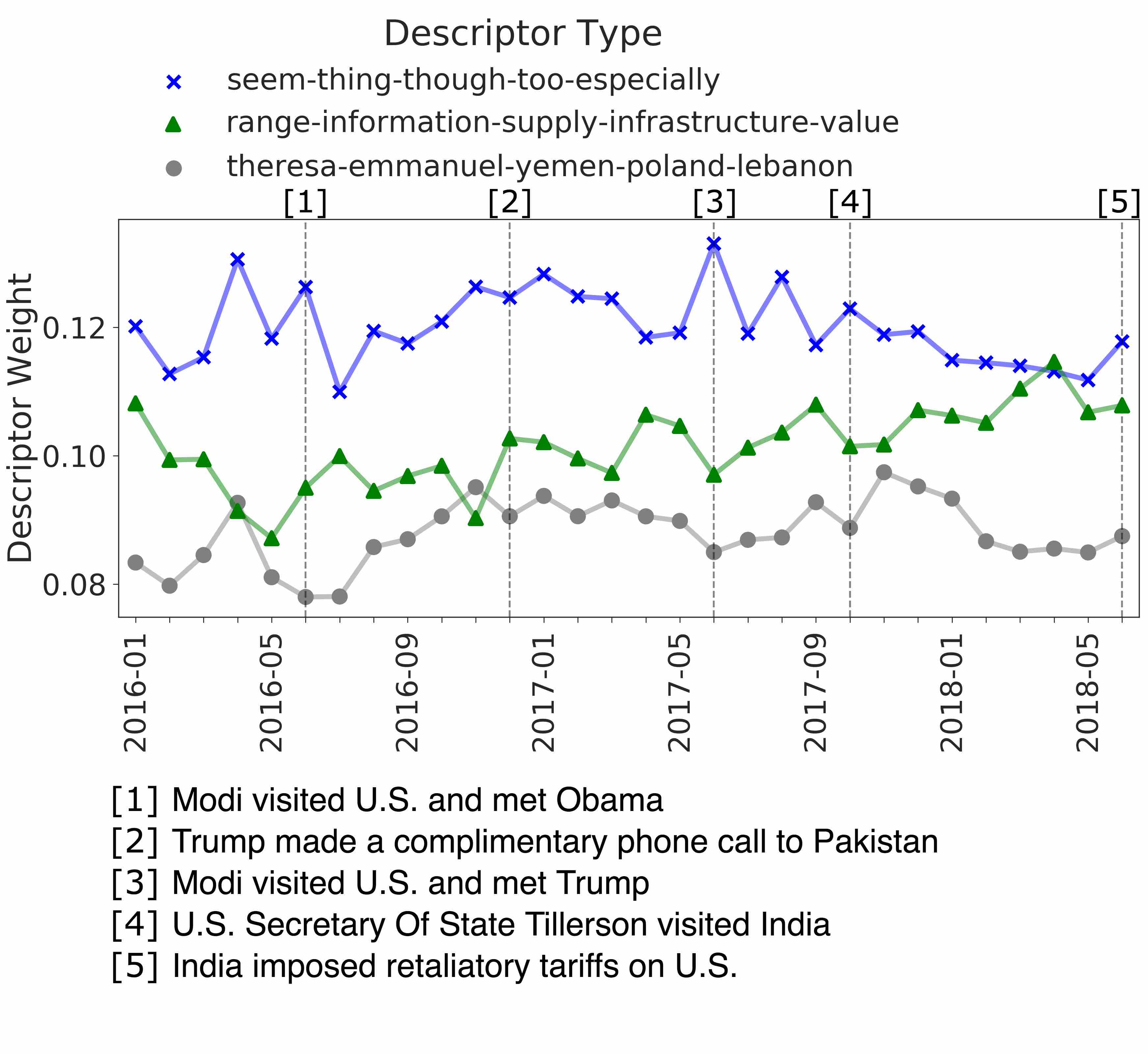}
    \caption{US-India's relation trends by \rmn.}
    \label{fig:UsIn_trend_mohit}
    \end{subfigure}
    \caption{
    Temporal trends of top three relations between U.S. and India based on \modelname (Figure~\ref{fig:UsIn_trend_ours}), in comparison with results from \rmn (Figure~\ref{fig:UsIn_trend_mohit}).
    }
    \label{fig:UsIn_trends}
\end{figure*}

\begin{figure*}[t]
    \begin{subfigure}[h]{0.47\textwidth}
    \includegraphics[width=\textwidth]{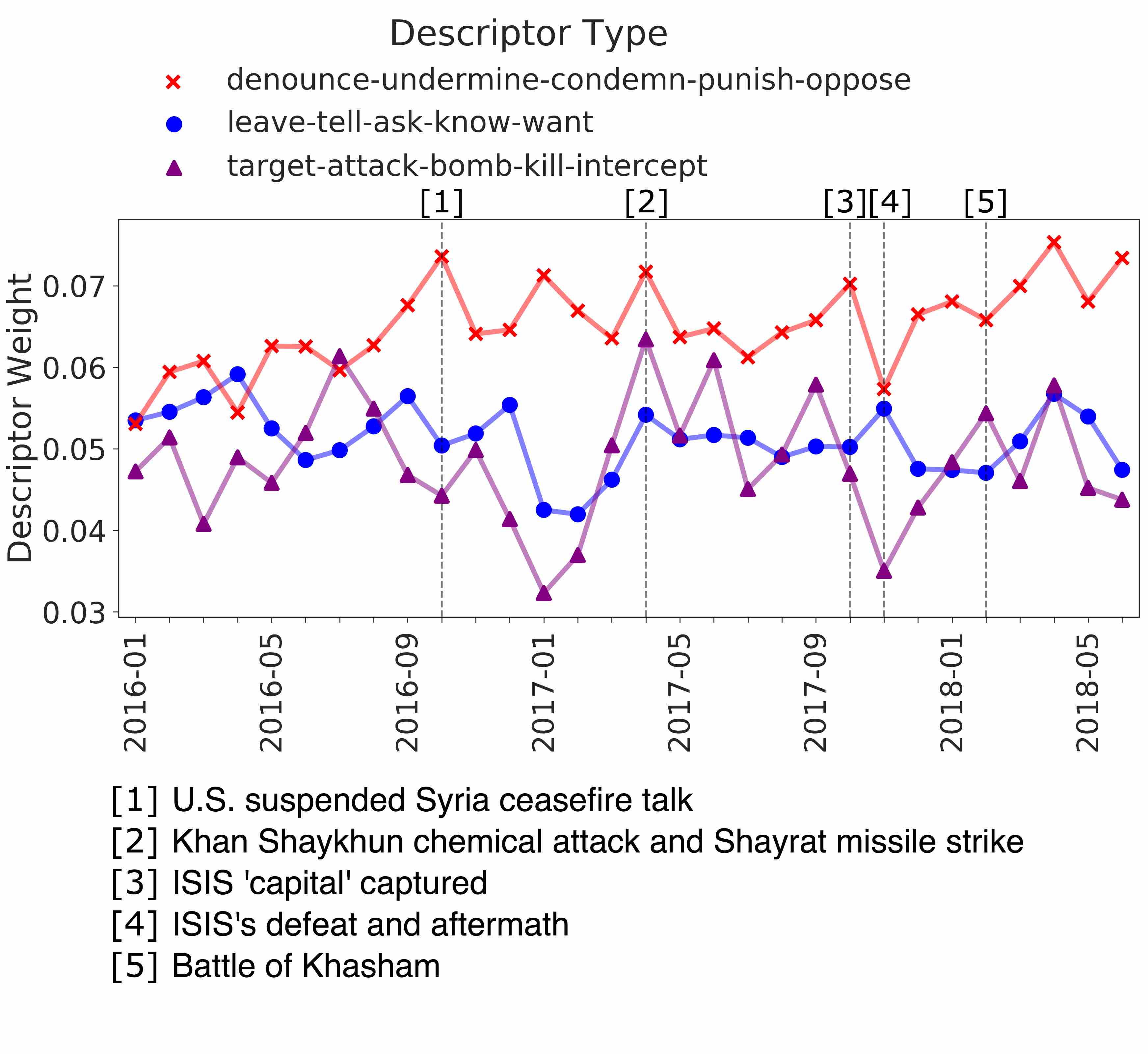}
    \caption{US-Syria's relation trends by \modelname.}
    \label{fig:UsSy_trend_ours}
    \end{subfigure}
    \hfill
    \begin{subfigure}[h]{0.47\textwidth}
    \includegraphics[width=\textwidth]{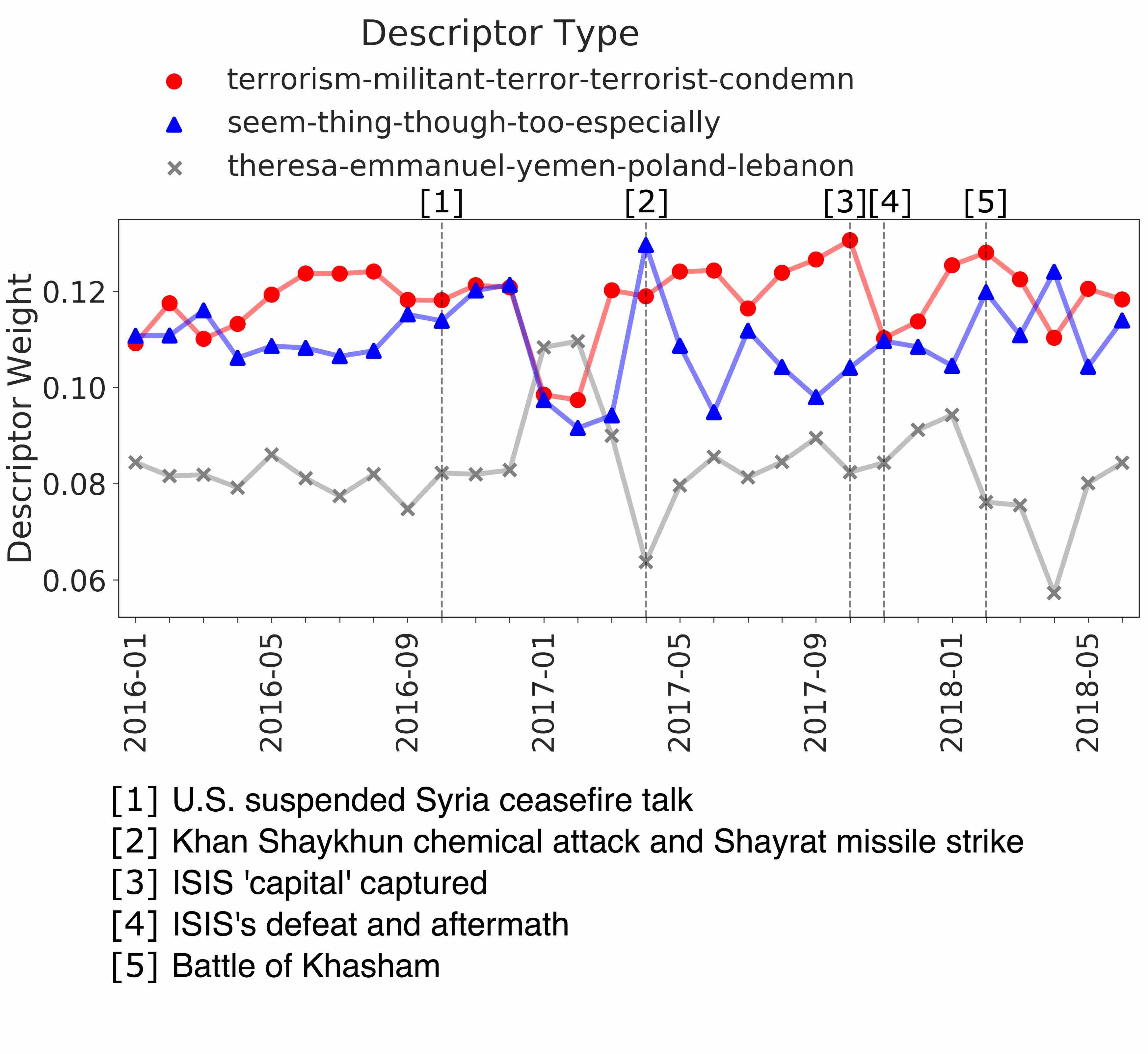}
    \caption{US-Syria's relation trends by \rmn.}
    \label{fig:UsSy_trend_mohit}
    \end{subfigure}
    \caption{
    Temporal trends of top three relations between U.S. and Syria based on \modelname (Figure~\ref{fig:UsSy_trend_ours}), in comparison with results from \rmn (Figure~\ref{fig:UsSy_trend_mohit}).
    }
    \label{fig:UsSy_trends}
\end{figure*}

See Figure~\ref{fig:UsRu_trends}, Figure~\ref{fig:UsIn_trends}, and Figure~\ref{fig:UsSy_trends} for three more examples of temporal relation trends between nation pairs. 

\subsection*{Additional Influential Background Words:US-Russia, US-India, US-Syria}
\begin{figure}[t]
  \centering
  \includegraphics[width=0.47\textwidth]{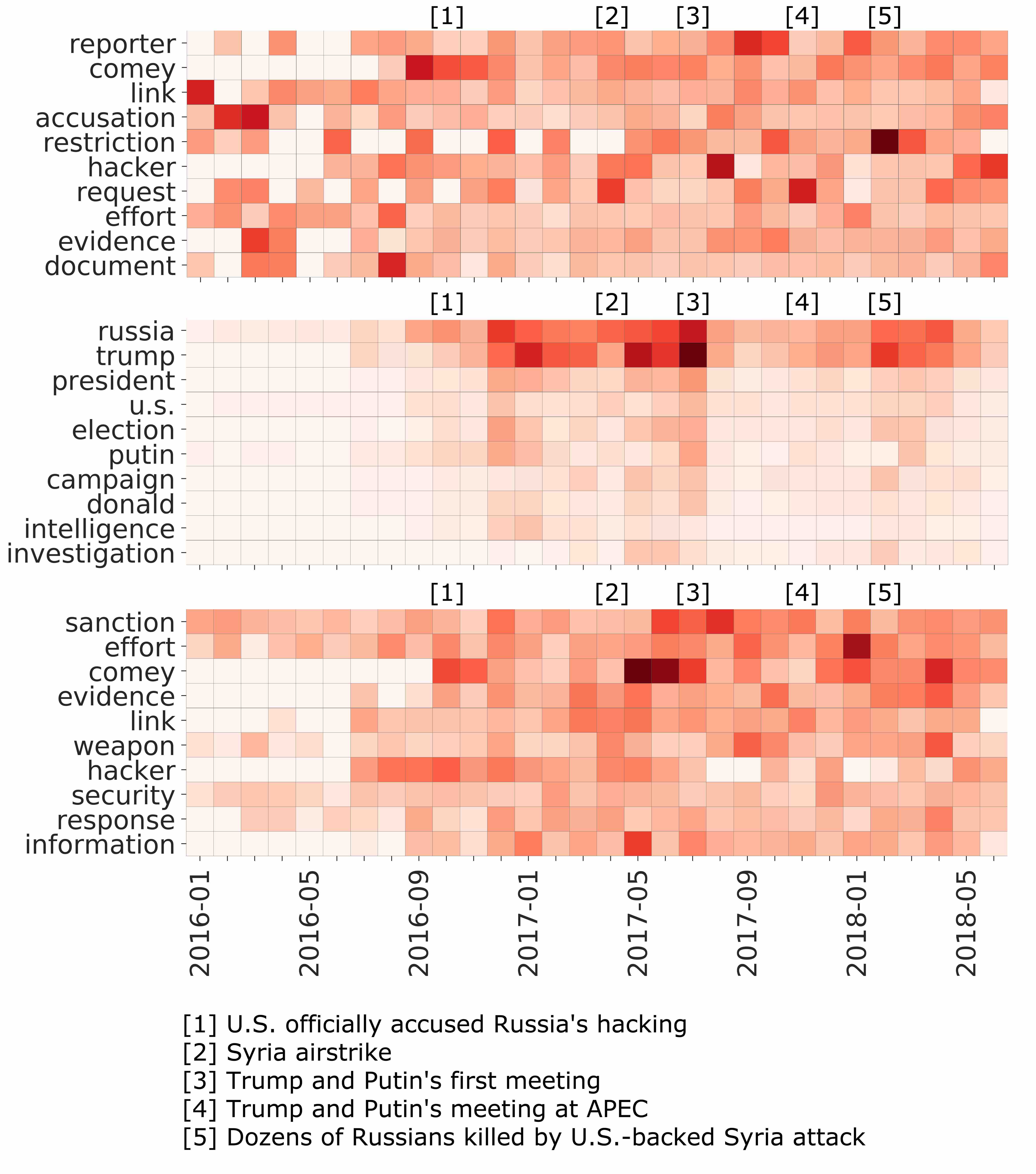}
  \caption{Influential words for US-Russia's ``denounce" relation derived from three approaches. \textit{Top}: Showing word's average attention score in all ``denounce'' articles each month. \textit{Middle}: Showing word's appearance frequency in all ``denounce'' articles each month. \textit{Bottom}: Showing word's average attention score multiplied by $\log(\text{appearance})$ in each month. All figures are normalized by the global maximum score in the figure.
  }
  \label{fig:aw_UsRu}
\end{figure}

\begin{figure}[t]
  \centering
  \includegraphics[width=0.47\textwidth]{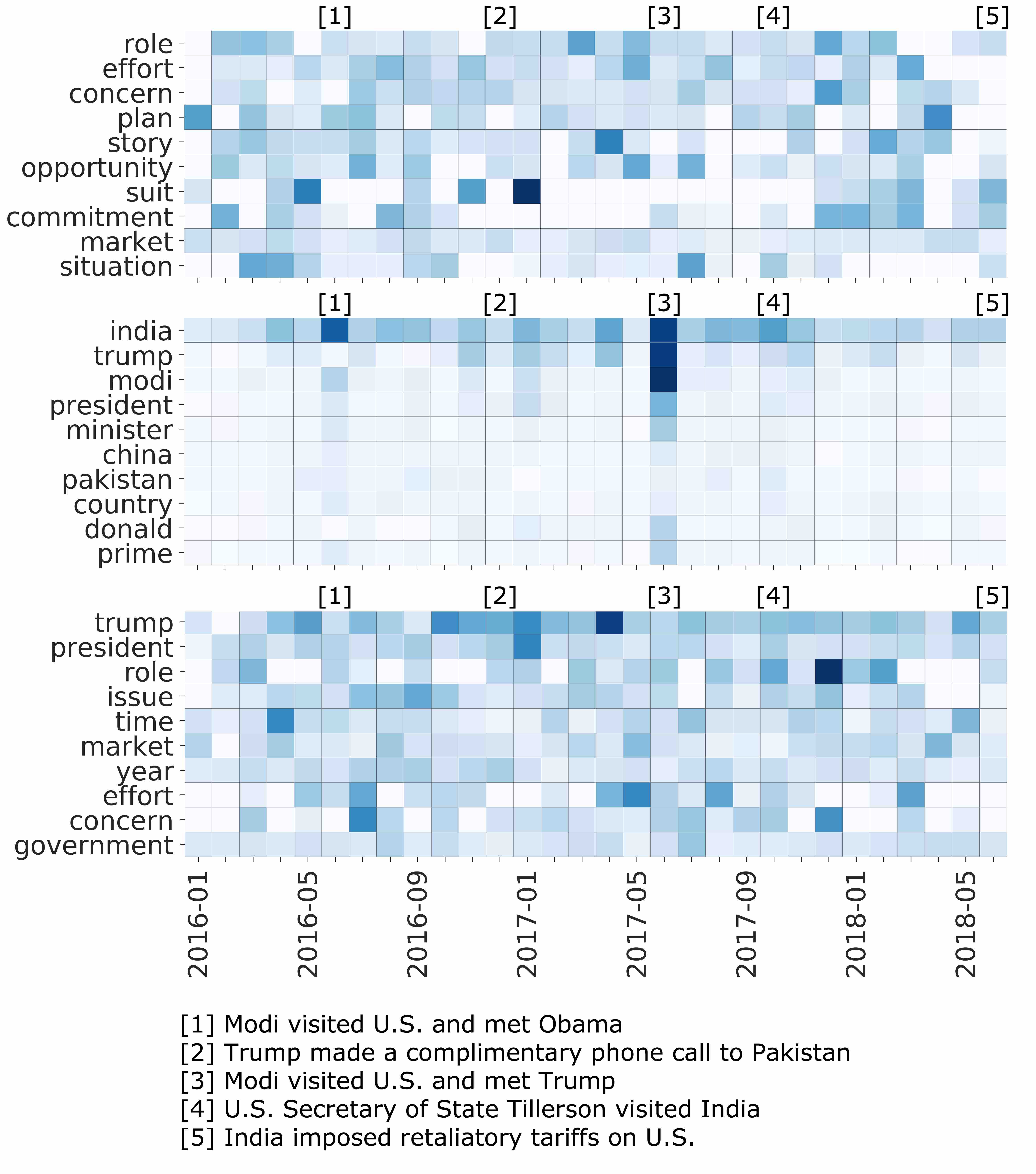}
  \caption{Influential words for US-India's ``leave" relation derived from three approaches. \textit{Top}: Showing word's average attention score in all ``leave'' articles each month. \textit{Middle}: Showing word's appearance frequency in all ``leave'' articles each month. \textit{Bottom}: Showing word's average attention score multiplied by $\log(\text{appearance})$ in each month. All figures are normalized by the global maximum score in the figure.
  }
  \label{fig:aw_UsIn}
\end{figure}

\begin{figure*}[t]
  \centering
  \includegraphics[width=0.47\textwidth]{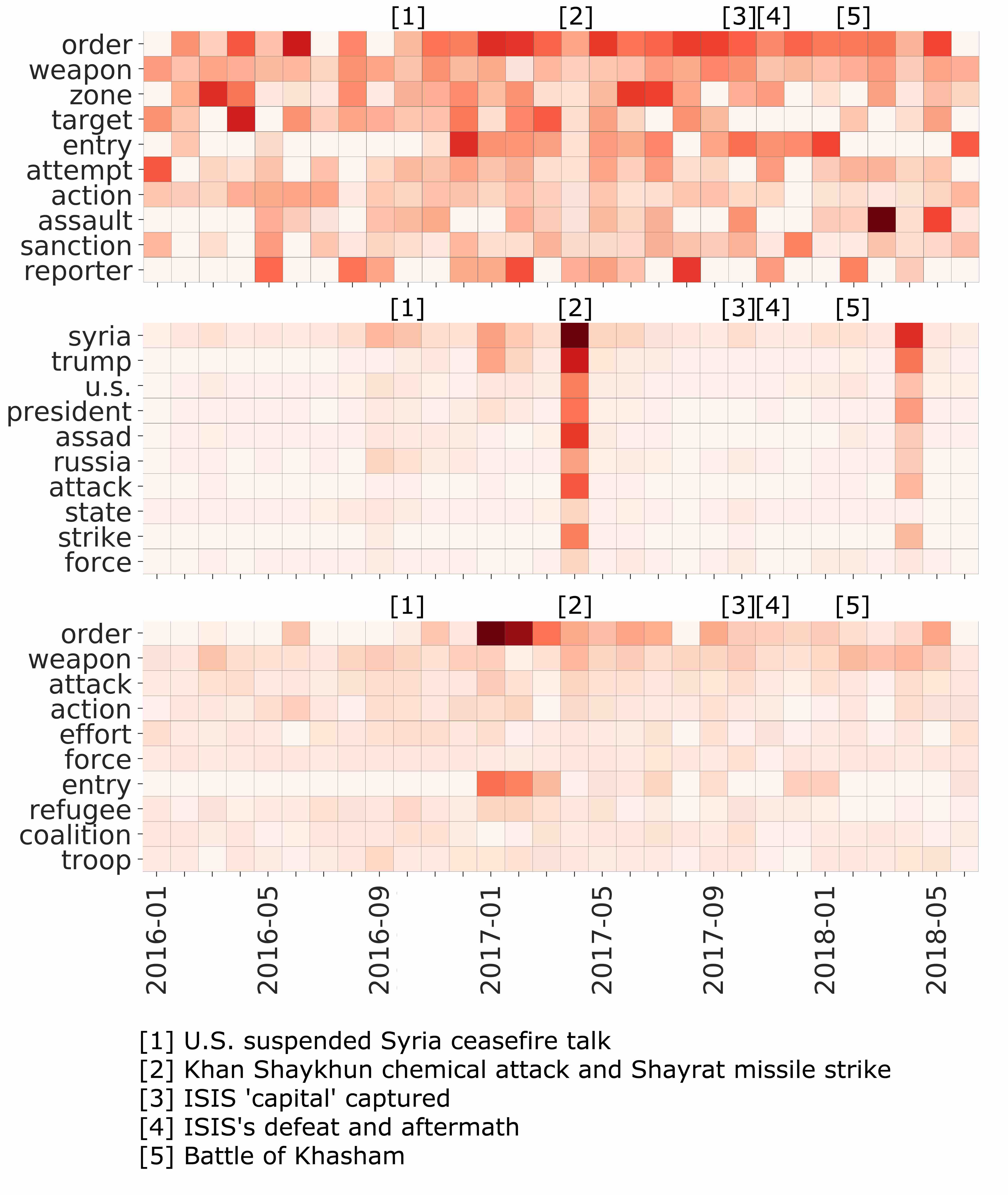}
  \caption{Influential words for US-Syria's ``denounce" relation derived from three approaches. \textit{Top}: Showing word's average attention score in all ``denounce'' articles each month. \textit{Middle}: Showing word's appearance frequency in all ``denounce'' articles each month. \textit{Bottom}: Showing word's average attention score multiplied by $\log(\text{appearance})$ in each month. All figures are normalized by the global maximum score in the figure.
  }
  \label{fig:aw_UsSy}
\end{figure*}

In Figure~\ref{fig:aw_UsRu}, Figure~\ref{fig:aw_UsIn}, and Figure~\ref{fig:aw_UsSy}, we further show the corresponding attended background words for the top relation descriptor in each of the three nation pair relations shown in the previous section. 

\subsection*{Robustness Check: Another Set of Key Events Annotations}
\begin{figure}[h]
    \centering
  \includegraphics[width=0.45\textwidth]{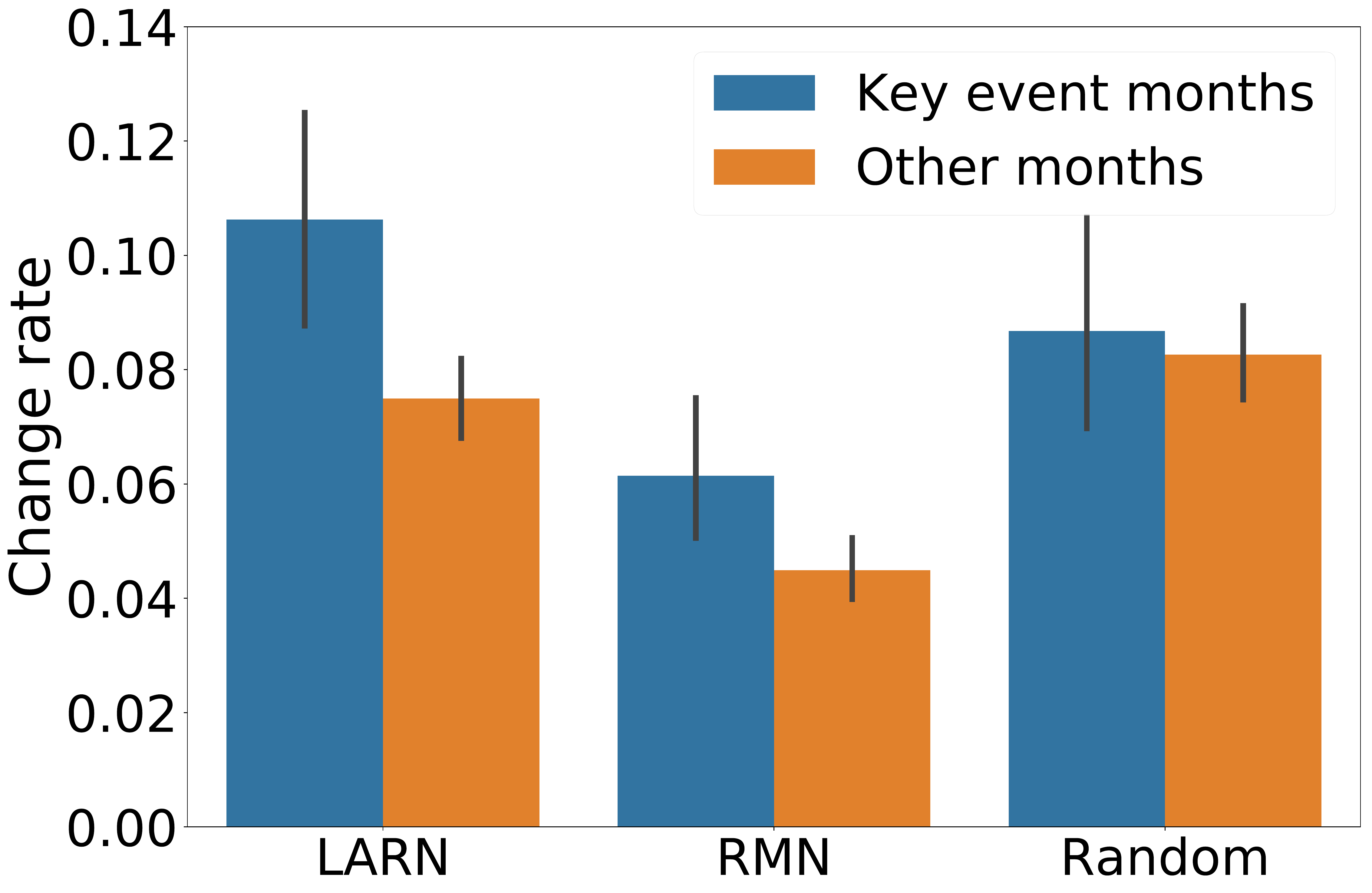}
  \caption{Change rate at key event months vs. other months using an additional set of key events annotations. Our model demonstrates a clearer relative difference between key event months and other months (41.74\%) than \rmn (36.75\%). For the random baseline, we randomly choose the same number of key events and we observe no differences as expected.
  }\label{fig:cpa_annot_2}
\end{figure}

Apart from the key events annotated by the first author, which were used in the change rate analysis in the main paper, the third author also did an independent annotation of the key events as a robustness check. The details of this annotation could found in Table~\ref{tab:robustness_check_events}. Figure~\ref{fig:cpa_annot_2} shows the change rate evaluation with respect to this annotation. A similar trend holds for both sets of annotations.

\subsection*{A Simpler Baseline: Term-frequency Trend of Verbal Predicates}
\begin{figure*}[t]
    \begin{subfigure}[h]{0.47\textwidth}
    \includegraphics[width=\textwidth]{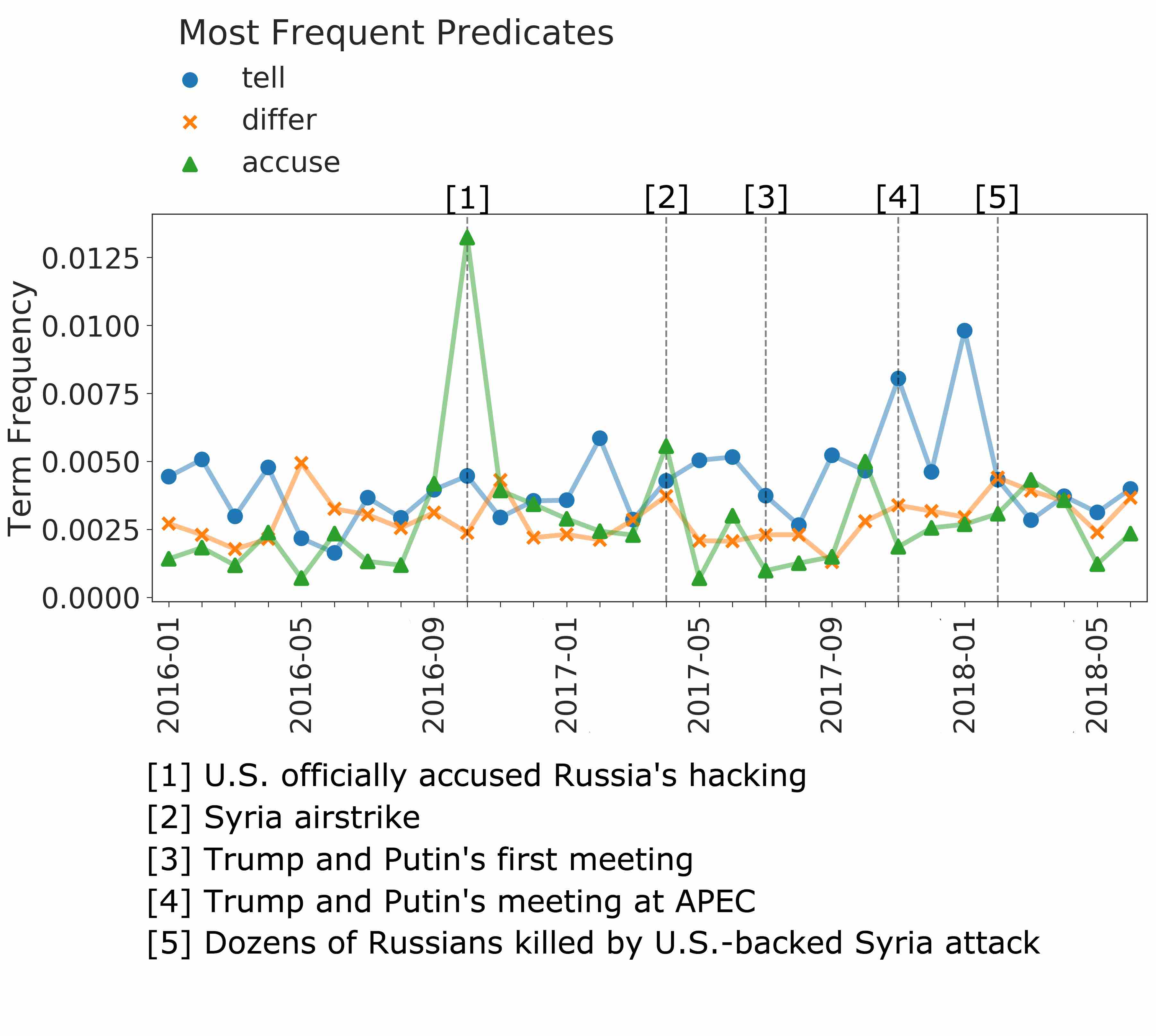}
    \caption{US-Russia's relation}
    \label{fig:UsRu_tf}
    \end{subfigure}
    \hfill
    \begin{subfigure}[h]{0.47\textwidth}
    \includegraphics[width=\textwidth]{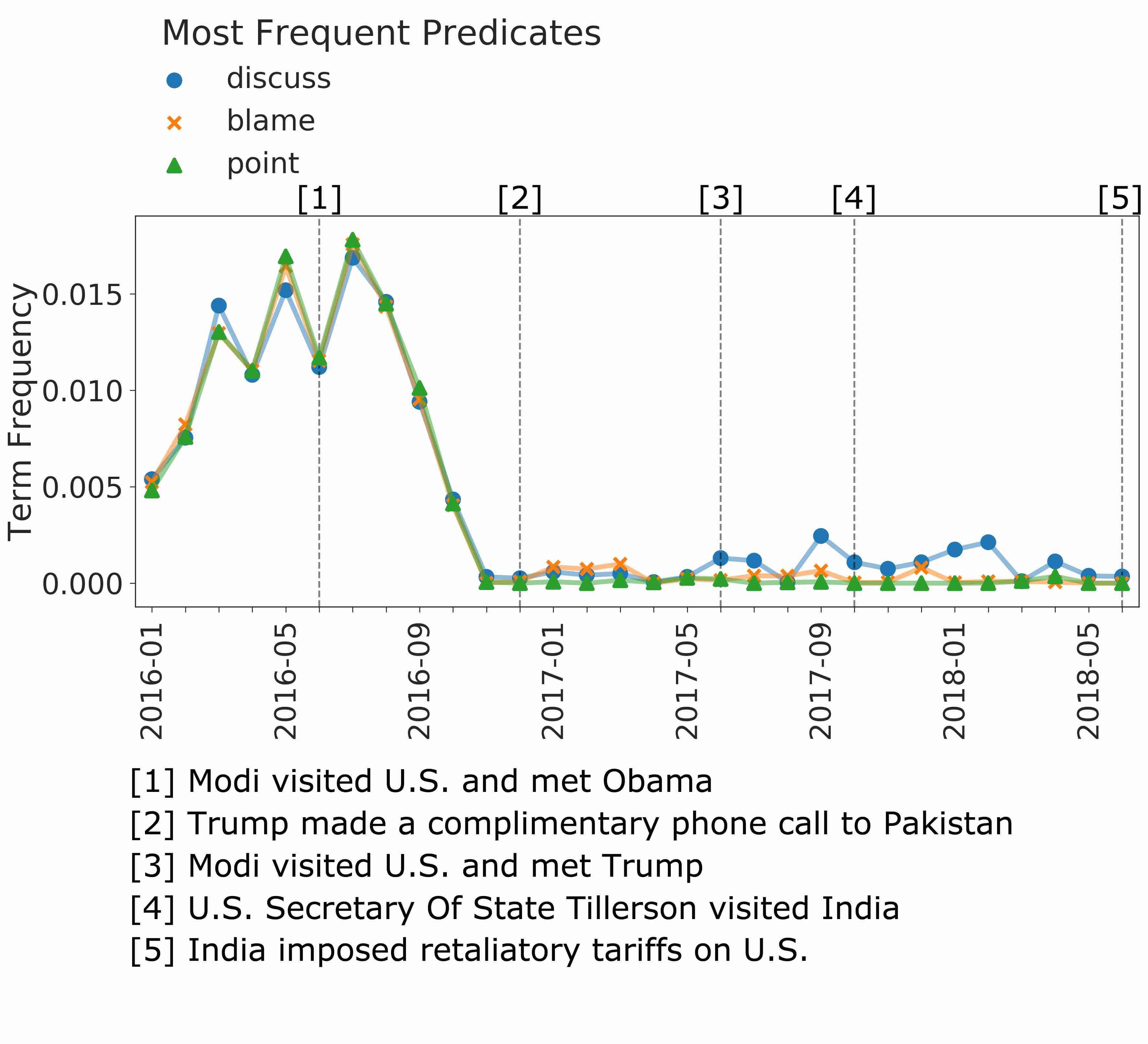}
    \caption{US-India's relation}
    \label{fig:UsIn_tf}
    \end{subfigure}
    \caption{
    Temporal trend examples by the predicate term-frequency baseline.
    }
    \label{fig:simple_baseline_tt}
\end{figure*}

In addition to the comparison with \rmn, we also checked a simpler baseline by replacing the descriptor weights with term-frequency values of each verbal predicates for each document.
Specifically, the term-frequency of a predicate $p$ in document $d$ is $\dfrac{count(p)}{length(d)}$.
Note that {\em document} refers to all sentences where a nation pair is mentioned.
We show some temporal trend examples in Figure~\ref{fig:simple_baseline_tt} and also the change rate evaluation result in Table~\ref{tab:simple_baseline_cpa}. Note that the scales of these results are different from \modelname and \rmn, since the predicate words are much more sparse than the 30 descriptors used in \modelname and \rmn.

\begin{table}[h]
\small
\begin{center}
\begin{tabular}{l|r|r|r}
\toprule
& \multicolumn{2}{c|}{Change rate} & \multirow{2}{*}{$\Delta$} \\
 & key months & other months  &   \\
\midrule
\modelname & 12.44 \% & 7.45 \% & 66.91 \%\\
\rmn & 6.81 \% & 4.53 \% & 50.07 \%\\
\midrule
Predicate TF & 69.86 \% & 62.36 \% & 12.02 \%\\
\bottomrule
\end{tabular} 
\end{center}
\caption{Change rate at key event months vs. other months with the predicate term-frequency baseline, compared to \modelname and \rmn.}
\label{tab:simple_baseline_cpa}
\end{table}

\clearpage

\begin{table*}
\small
\begin{center}
\renewcommand{\arraystretch}{1.2}
\begin{tabular}{lp{12cm}}
\toprule
Nation pair & Key events\\
\midrule
U.S.-China
 & (2016-12) \href{https://en.wikipedia.org/wiki/Trump–Tsai_call}{Trump made phone call to Taiwan's leader};
 (2017-04) \href{https://www.theguardian.com/us-news/2017/apr/06/trump-china-meeting-xi-jinping-mar-a-lago}{Xi visited U.S.};
 (2017-08) \href{http://www.businessinsider.com/us-begins-section-301-investigation-2017-8}{Section 301 investigations on China};
 (2017-11) \href{https://www.nytimes.com/2017/11/07/business/trump-china-trade.html}{Trump visited China};
 (2018-03) \href{https://www.nytimes.com/2018/03/22/us/politics/trump-will-hit-china-with-trade-measures-as-white-house-exempts-allies-from-tariffs.html}{Trump started issuing a series of tariffs}.\\
U.S.-Russia & (2016-10) \href{https://www.theguardian.com/technology/2016/oct/07/us-russia-dnc-hack-interfering-presidential-election}{U.S. officially accused Russia's hacking};
 (2017-04) \href{https://www.theguardian.com/world/2017/apr/07/us-airstrikes-syria-russian-american-relations-vladimir-putin}{Syria airstrike};
 (2017-07) \href{https://www.theatlantic.com/news/archive/2017/07/trump-putin/532899/}{Trump and Putin's first meeting};
 (2017-11) \href{https://www.cnn.com/2017/11/09/politics/donald-trump-vladimir-putin-vietnam/index.html}{Trump and Putin's meeting at APEC};
 (2018-02) \href{https://www.nytimes.com/2018/02/13/world/europe/russia-syria-dead.html}{Dozens of Russians killed by U.S.-backed Syria attack}.\\
U.S.-Syria & (2016-10) \href{https://www.reuters.com/article/us-mideast-crisis-usa-russia-idUSKCN1231X3}{U.S. suspended Syria ceasefire talk};
 (2017-04) \href{https://en.wikipedia.org/wiki/2017_Shayrat_missile_strike}{Khan Shaykhun chemical attack and Shayrat missile strike};
 (2017-10)
 \href{https://www.nytimes.com/2017/10/17/world/middleeast/isis-syria-raqqa.html}{ISIS 'capital' captured};
 (2017-11)
 \href{https://news.antiwar.com/2017/11/17/pentagon-isis-defeated-but-us-will-stay-in-syria/}{ISIS's defeat and aftermath};
 (2018-02)
 \href{https://en.wikipedia.org/wiki/Battle_of_Khasham}{Battle of Khasham}.\\
U.S.-U.K. & (2016-06) \href{https://www.bbc.com/news/36622711}{U.K. Brexit vote};
 (2017-01)
 \href{https://www.cnn.com/2017/01/26/politics/donald-trump-theresa-may-white-house-visit/index.html}{Therasa May visited U.S.};
 (2017-12)
 \href{https://www.csmonitor.com/World/Europe/2017/1207/In-Trump-era-US-UK-special-relationship-faces-and-causes-new-trials}{Trump set a controversial visit to U.K.} \\
U.S.-Canada
 & (2017-01)
 \href{https://www.bbc.com/news/world-us-canada-38713227}{Trump said Nafta renegotiation to be started};
 (2017-04)
 \href{https://www.nytimes.com/2017/04/24/us/politics/lumber-tariff-canada-trump.html}{Trump imposed tariff on Canadian lumber};
 (2018-06)
 \href{https://www.cnbc.com/2018/06/29/canada-makes-retaliatory-tariffs-official-we-will-not-back-down.html}{Canada fought back with retaliatory tariff on U.S. products}.\\
U.S.-India
 & (2016-06)
 \href{https://www.c-span.org/video/?410278-2/president-obama-meets-prime-minister-modi-india}{Modi visited U.S. and met Obama};
 (2016-12)
 \href{https://www.bbc.com/news/world-asia-38165878}{Trump made a complimentary phone call to Pakistan};
 (2017-06)
 \href{https://www.c-span.org/video/?430524-4/us-india-relations}{Modi visited U.S. and met Trump};
 (2017-10)
 \href{https://www.npr.org/sections/parallels/2017/10/26/560224471/tillerson-visit-highlights-indias-evolving-relationship-with-u-s}{U.S. Secretary Of State Tillerson visited India};
 (2018-06)
 \href{https://www.washingtonpost.com/world/india-imposes-retaliatory-tariffs-on-us-as-global-trade-war-widens/2018/06/21/7c3a016b-1de0-497a-9635-a522bc55810a_story.html?noredirect=on&utm_term=.6ba2cd6411ed}{India imposed retaliatory tariffs on U.S.}\\
U.S.-Japan
 & (2016-05)
 \href{https://www.nytimes.com/2016/05/28/world/asia/text-of-president-obamas-speech-in-hiroshima-japan.html}{Obama gave memorial speech at Hiroshima with Japanese PM Abe};
 (2016-12)
 \href{https://www.nytimes.com/2016/12/05/world/asia/shinzo-abe-pearl-harbor-japan.html}{Abe visited Pearl Harbor};
 (2017-02)
 \href{https://www.washingtonpost.com/news/monkey-cage/wp/2017/02/13/did-trump-and-abe-just-launch-a-new-chapter-in-u-s-japan-relations/?utm_term=.f4e2a17f96b6}{Abe visited Washington and met Trump};
 (2017-11)
 \href{https://www.whitehouse.gov/briefings-statements/president-donald-j-trumps-summit-meeting-prime-minister-shinzo-abe-japan/}{Trump visited Japan and met Abe};
 (2018-03)
 \href{https://www.cnbc.com/2018/03/29/japan-wants-its-own-bilateral-summit-with-north-korea.html}{Trump accepted North Korea's invitation for direct nuclear talks}.\\
China-India
 & (2016-04)
 \href{https://thediplomat.com/2016/04/foreign-ministers-of-russia-india-china-meet-in-moscow/}{Minister of Foreign Affairs meeting};
 (2016-11)
 \href{https://thediplomat.com/2016/11/china-india-hold-joint-military-drill/}{China and India's joint military drill};
 (2017-02)
 \href{https://www.armyrecognition.com/weapons_defence_industry_military_technology_uk/india_to_develop_new_variant_of_brahmos_missile.html}{India to develop a new missile};
 (2017-05)
 \href{https://www.thehindu.com/news/international/india-unlikely-to-participate-in-chinas-belt-and-road-forum/article18445908.ece}{India refused to attend Belt and Road Summit};
 (2017-06)
 \href{https://en.wikipedia.org/wiki/2017_China–India_border_standoff}{Doklam border standoff started};
 (2017-08)
 \href{https://en.wikipedia.org/wiki/2017_China–India_border_standoff}{Doklam border standoff ended};
 (2017-11)
 \href{http://mea.gov.in/press-releases.htm?dtl/29122/IndiaChina_WMCC_Meeting_November_17_2017}{China and India's WMCC meeting};
 (2018-03) 
 \href{http://www.xinhuanet.com/english/2018-03/27/c_137068871.htm}{China and India to boost trade}.\\
\bottomrule
\end{tabular} 
\end{center}
\caption{Key events annotations and references.}
\label{tab:events}
\end{table*}

\clearpage

\clearpage

\begin{table*}
\small
\begin{center}
\renewcommand{\arraystretch}{1.2}
\begin{tabular}{lp{12cm}}
\toprule
Nation pair & Key events\\
\midrule
U.S.-China
 & 
(2016-02) China Sends Missiles to Contested South China Sea Island; 
(2016-06) Xi urges caution over THAAD deployment in South Korea; 
(2016-12) Trump-Tsai call; 
(2017-02) Trump affirms One China policy; 
(2017-04) Xi visits US; 
(2017-11) Trump visited Beijing; 
(2018-03) tariff targets China; 
(2018-04) China retaliates. 
\\
U.S.-Russia
 & 
(2016-02) U.S.-Russian deal on a ``cessation of hostilities''; 
(2016-09) Russia and the U.S. announce joint peace plan for Syria; 
(2016-10) Russian interference with election; 
(2017-01) The U.S. Office of the Director of National Intelligence releases a declassified version of the Intelligence Community’s assessment that ``Putin ordered an influence campaign in 2016 aimed at the U.S. presidential election''; 
(2017-04) US airstrike Syria; 
(2017-05) Comey fired and Trump met Russian ambassadors; 
(2017-07) Trump met Putin at G20 summit; 
(2018-04) Missile strikes against Syria.
\\
U.S.-Syria
 & 
(2016-02) US and Russia agree to enforce new Syria ceasefire; 
(2016-09) Another ceasefire agreement broke in the same month; 
(2016-11) Senior Chief Petty Officer Scott Cooper Dayton, 42, of Woodbridge, Virginia, is killed in an improvised explosive device blast near the northern Syrian town of Ayn Issa, becoming the first American casualty in combat in the fight against IS in Syria.; 
(2017-01) Trump orders ban on Syrian refugees; 
(2017-04) Trump orders strikes against Syria in response to chemical attack; 
(2017-07) Another ceasefire agreement that involved united states; 
(2018-03) U.S. Master Sgt. Jonathan J. Dunbar is killed by a roadside bomb attack in Syria alongside a British serviceman; 
(2018-04) The US, The U.K. and France carried out missile strikes against Assad's compounds in response to the Douma chemical attack.
\\
U.S.-U.K.
 & 
(2016-04) Obama and Cameron joint news conference; 
(2017-01) May visit Trump; 
(2017-02) Parliament debate trump visit; 
(2017-09) May visit US for UN general assembly; 
(2018-04) Missile strikes in Syria.
\\
U.S.-Canada
 & 
(2016-03) Trudeau visited the White House for an official visit and state dinner on March 10, 2016; 
(2017-02) Prime Minister Trudeau and President Trump formally met for the first time at the White House on February 13, 2017; 
(2017-06) The Trudeau government announced that Canada would continue to support coalition operations; 
(2018-03) President Donald J. Trump announced he will apply across-the-board tariffs, or import taxes, on steel and aluminum; 
(2018-05) Canada, Mexico, and the EU became subject to the steel and aluminium tariffs later in an announcement on May 31, 2018; 
(2018-06) Trump comments at G7 summit.
\\
U.S.-India
 & 
(2016-02) The Obama administration notified the US Congress that it intended to provide Pakistan eight nuclear-capable F-16 fighters and assorted military goods; 
(2016-06) Modi visits Obama; 
(2016-08) U.S., India sign military logistics agreement; 
(2017-06) Modi visits Trump; 
(2017-09) Mattis visits India; 
(2017-10) Secretary of State Rex Tillerson visits India; 
(2018-03) The India-US 2 + 2 Dialogue postponed.
\\
U.S.-Japan
 & 
(2016-05) Abe meets Obama; 
(2016-07) Abe, U.S. commander agree to carry out defense guidelines in steady manner; 
(2016-11) Abe meets Trump in New York; 
(2016-12) U.S. Returns Some Okinawa Land to Japan, but Unease Endures; Abe visits Pearl Harbor; 
(2017-01) Trump abandons TPP; 
(2017-02) Abe and Trump presented a united front on dealing with Pyongyang's nuclear weapon test and multiple missile launches; 
(2017-04) Japan, U.S. hold missile defense drill in Sea of Japan; 
(2017-09) Japan, U.S., India vow to work together on strategic port development as China flexes clout; 
(2017-11) Trump visits Japan; 
(2018-02) Pence visits Japan; 
(2018-04) Abe visits US. 
\\
China-India
 & 
(2016-06) Tashkent on the sidelines of the Shanghai Cooperation Organization Summit; 
(2016-09) G20 and East Asia Summit; 
(2016-10) Modi meets Chinese president Xi Jinping on the sidelines of the Goa BRICS Summit; 
(2017-05) India’s decision to boycott the Belt and Road Initiative (BRI) summit held in Beijing in May; 
(2017-06) Chinese troops with construction vehicles and road-building equipment began extending an existing road southward in Doklam, a territory which is claimed by both China as well as India's ally Bhutan.; 
(2017-08) China and India reached a consensus to put an end to the border stand-off; 
(2017-09) Xi Jinping Meets with Prime Minister Narendra Modi of India; 
(2017-11) India joins QUAD; 
(2018-03) China-India border affairs meeting held in New Delhi; 
(2018-06) Xi meets Modi.
\\
\bottomrule
\end{tabular} 
\end{center}
\caption{Extra key events annotations for robustness check.}
\label{tab:robustness_check_events}
\end{table*}

\end{document}